# How Much Is Too Much? Adaptive, Context-Aware Risk Detection in Naturalistic Driving

Amir Hossein Kalantari, Eleonora Papadimitriou, Arkady Zgonnikov, and Amir Pooyan Afghari

***Abstract*—Reliable risk identification based on driver behavior data underpins real-time safety feedback, fleet risk management, and evaluation of driver-assist systems. While naturalistic driving studies have become foundational for providing real-world driver behavior data, the existing frameworks for identifying risk based on such data have two fundamental limitations: (i) they rely on predefined time windows and fixed thresholds to disentangle risky and normal driving behavior, and (ii) they assume behavior is stationary across drivers and time, ignoring heterogeneity and temporal drift. In practice, these limitations can lead to timing errors and miscalibration in alerts, weak generalization to new drivers/routes/conditions, and higher false-alarm and miss rates, undermining driver trust and reducing safety intervention effectiveness. To address this gap, we propose a unified, context-aware framework that adapts labels and models over time and across drivers via rolling windows, joint optimization, dynamic calibration, and model fusion, tailored for time-stamped kinematic data. The framework is tested using two safety indicators, speed-weighted headway and harsh driving events, and three models: Random Forest, XGBoost, and Deep Neural Network (DNN). Speed-weighted headway yielded more stable and context-sensitive classifications than harsh-event counts. XGBoost maintained consistent performance under changing thresholds, whereas DNN achieved higher recall at lower thresholds but with greater variability across trials. The ensemble aggregated signals from multiple models into a single risk decision, balancing responsiveness to risky behavior with control of false alerts. Overall, the framework shows promise for adaptive, context-aware risk detection that can enhance real-time safety feedback and support driver-focused interventions in intelligent transportation systems.**

## I. INTRODUCTION

TRAFFIC conflict data have been widely used for proactively analyzing road safety [1]. Conflicts are more frequent than crashes and may be used as their precursors, providing a unique opportunity for predicting crashes before they actually occur. A traffic conflict is defined as an observable situation in which two or more road users approach each other in space and time to such an extent that there is a risk of collision if their speed and direction of movement remain unchanged [2]. Numerous traffic conflict studies have adopted this definition, especially in the context of video analytics. In this approach, the trajectories of road users are first extracted from aerial videos, and their spatial and temporal proximity is then calculated to define conflicts [3], [4], [5]. Critical and serious conflicts are then identified by applying thresholds on surrogate safety measures (SSMs) such as time to collision (TTC), post encroachment time (PET), and Deceleration Rate to Avoid a Crash (DRAC) among others [1], [6]. These conflicts have been shown to correlate well with crashes [7], [8], [9].

Although aerial videos and infrastructure-based methods such as inductive loop detectors provide valuable aggregate insights into traffic conflicts, they are less effective in capturing individual-level behavioral nuances or monitoring driver behavior over time [5], [10]. They also fall short in linking conflicts to the psychosocial and demographic characteristics of individual road users. To address these gaps, naturalistic driving studies employing instrumented vehicles can be used, offering detailed longitudinal records of individual drivers that are particularly well-suited for analyzing risky driving behaviors.

Naturalistic driving data have been widely used in previous studies for modeling driving behavior and risk [11]. Within this stream, time headway and harsh events (e.g., abrupt braking/acceleration) act as proximal precursors of conflicts: headway governs closure time and reaction margins, whereas harsh deceleration reflects evasive responses when margins are violated [12], [13], [14]. These indicators are helpful in depicting unusual driver behavior (such as inattention) and can be interpreted as potential danger [15], [16]. This is particularly true in specific traffic contexts such as car following maneuvers. Research has shown that the likelihood of a crash remains low as long as the time headway of the following driver does not fall below their reaction time [17]. Nonetheless, shorter headways (usually less than two seconds) can generally increase the risk of rear-end collisions, as they leave drivers with less time to react to sudden changes in traffic conditions [18]. In addition, there is a relationship between an intense, jerky driving style and short headways, as it requires the driver to closely monitor and precisely respond to the movements of the vehicle ahead; failure to do so may result in either an increased headway which is not optimal for the drivers or, more critically, a collision [19]. On the other hand, harsh driving events, such as harsh braking or

This study was conducted with funding from the European Union's Horizon 2020 i-DREAMS project (Project No. 814761), supported by the European Commission under the MG-2-1-2018 Research and Innovation Action (RIA) programme.

The first and fourth authors are with the Safety and Security Science Section, Department of Values, Technology and Innovation, Faculty of Technology, Policy and Management, Delft University of Technology, 2628 BX Delft, The Netherlands. The second author is with the Department of Transportation Planning and Engineering, National Technical University of Athens, 5 Iroon Polytechneiou St., GR-15773 Athens, Greece; at the time of the research she was with the Safety and Security Science Section, Delft University of Technology. The third author is with the Department of Cognitive Robotics, Delft University of Technology, 2628 Delft, The Netherlands. Corresponding author: Amir Hossein Kalantari (e-mail: a.h.kalantari@tudelft.nl).

acceleration, have been found to strongly correlate with individual road user behavior and spatial crash counts. The occurrence of harsh deceleration events is closely associated with an increased likelihood of crashes among individual drivers [20], with areas experiencing higher counts of harsh braking and acceleration also showing a greater number of collisions [21]. Overall, time headway and harsh events are practical conflict-proximal indicators: they are measurable, interpretable, and consistently associated with crash outcomes [22], [23]. In naturalistic driving studies, they capture pre-conflict conditions and evasive responses that often precede or escalate into critical conflicts, providing a naturalistic complement to TTC/PET-based inferences derived from trajectory video.

Working with naturalistic driving data, however, poses a fundamental challenge: transforming vast, heterogeneous sensor streams into a coherent structure that can reveal meaningful behavioral patterns. These datasets often contain millions of records collected at different sampling rates and with inevitable gaps, making direct analysis difficult. Although timestamp-based methods help mitigate some of these discrepancies [24], real-time streaming and missing data remain significant obstacles. A widely used approach to make such data analytically tractable is to aggregate observations within sliding time windows. The window length sets the temporal scale of risk assessment: short windows capture fine-grained fluctuations but may exaggerate noise, while longer windows reveal broader patterns yet risk obscuring brief critical events. The appropriate duration depends on traffic context, driver behavior, vehicle dynamics, and environmental conditions, requiring an optimized balance between data efficiency and analytical objectives.

Defining risky driving behavior within these adaptive windows presents a further challenge, as the thresholds used for classification are not universal but vary across contexts. Naturalistic studies show that harsh-braking events are highly sensitive to the chosen cut-off, with both event frequency and the relative ranking of drivers shifting when thresholds move, and with higher rates in urban settings compared to rural roads or motorways [25]. This context-dependence is not limited to harsh events. Similar variability is evident in following behavior, where headway distributions differ systematically with operating speed, visibility, and roadway type, meaning that no single “safe” headway applies across all environments [26]. In surrogate safety analyses, thresholds for TTC and PET likewise vary widely (TTC ≈ 1.5–4 s; PET ≈ 0–6 s), and the number of detected conflicts depends directly on the chosen value [27], [28]. Telematics-based studies further show that the frequency of harsh events is shaped by prevailing traffic flow and road geometry, underlining the importance of context-aware decision rules [29], [30]. Finally, driver-adaptive warning frameworks that adjust thresholds online (e.g., forward collision warning based on driver behavior) demonstrate feasibility and improved alignment to conditions [31]. Taken together, these findings highlight that thresholds, like window size, should be treated as scale- and context-dependent parameters rather than fixed constants.

This study addresses the above gaps by proposing a dynamic, data-driven framework for identifying risky driving behaviors from naturalistic data. It leverages a rolling time-window to capture context-specific variation in critical time headways and harsh events. State-of-the-art models are tuned with Bayesian optimization, and risk thresholds adapt over time rather than remaining fixed. The approach uses signals available in the dataset and employs calibration and validation to support reliable decisions. Additionally, post-hoc explainability (SHAP) provides insights into risky behaviors. The framework is tested and validated on i-DREAMS naturalistic driving dataset which is described in the next section.

## II. i-DREAMS Naturalistic driving study

The EU-funded Horizon 2020 i-DREAMS (intelligent Driver and Road Environment Assessment and Monitoring System) project aimed to develop, test, and validate a context-aware platform for promoting safe driving. The project involved studies conducted across five European countries: Greece, the United Kingdom, Portugal, Belgium, and Germany. It utilized naturalistic driving data for 250 drivers from these countries, resulting in a substantial dataset comprising 49,651 trips and 1,956,332 minutes of driving data. The experimental design of the i-DREAMS on-road study is structured into four consecutive phases:

- Phase 1: monitoring (baseline measurement)
- Phase 2: real-time intervention
- Phase 3: real-time intervention and post-trip feedback
- Phase 4: real-time intervention and post-trip feedback and gamification

Each vehicle was equipped with an OBD-II device compatible with standard protocols, supported by a Software Development Kit (SDK) for extracting vehicle kinematics and sensor data, along with a comprehensive set of Application Programming Interfaces (APIs) for integration with third-party systems. The OBD-II unit also featured 2G or 3G GSM/GPRS connectivity, enabling the transmission of sensor-recorded vehicle data to remote cloud servers. Data were transmitted automatically via the mobile network, without any need for user intervention.

Additionally, data were obtained from the Mobileye system [32], a dash camera and the Cardio gateway (CardioID Technologies, 2022). The Cardio Gateway captures vehicle dynamics metrics, including speed, acceleration, deceleration, and steering. Global Navigation Satellite System (GNSS) signals were also recorded as part of the multi-sensor data acquisition process. It is a sensor-based system connected to the Mobileye equipment via the vehicle’s CAN bus, and it supports data transmission through various communication technologies. Finally, smartphone-based telematics were also employed in conjunction with the aforementioned vision-based commercial systems and dash cameras, to collect naturalistic driving data.

The project considered background factors related to the driver, real-time physiological indicators associated with driving risk, and the complexity of driving tasks in order to assess whether a driver remained within the “safety tolerance zone” (STZ) during their daily trips. The concept of the STZ is based on Fuller’s Task Capability Interface Model [34] stating that loss of control occurs when the demand of a driving task outweighs the operator’s capability. The STZ comprises three phases: a normal driving phase, a danger phase and an avoidable accident phase. Within a transportation system, a driver can be viewed as a human operator (technology assisted) self-regulating control in either of the categories over vehicles in the context of crash avoidance. When a driver approached predefined safety thresholds, the aforementioned interventions, both real-time and post-trip, were triggered. The dataset included a wide range of variables, such as trip duration, distance travelled, speeding incidents, mobile phone usage while driving, harsh braking, harsh acceleration, harsh cornering (i.e., lateral deviation), time headway (continuous), headway levels, and warnings issued by advanced driver-assistance systems (ADAS), including Forward Collision Warning (FCW), Pedestrian Collision Warning (PCW), Lane Departure Warning (LDW), and Speed Limit Indication signals. It also captured contextual factors such as the start and end of forbidden overtaking zones, as well as physiological metrics like heart rate inter-beat interval (IBI) and fatigue warning [35], [36]. In addition, drivers’ socio-demographic characteristics, as well as their beliefs, attitudes, traffic accident and offence history, driving style, and other relevant factors, were collected through a comprehensive survey questionnaire.

For the present study, only the naturalistic driving data of Belgian drivers were used, comprising 52 participants enrolled in the experiment. Of these, 39 drivers participated consistently across all four phases, contributing a total of 7,163 trips and 147,337 minutes of driving data. Seventy percent of the drivers were male, and their ages ranged from 20 to 70 years, with a mean age of 44 years.

Table I shows the selected variables from the study for modeling purposes. For a detailed description of all variables and values in the i-Dreams experiment, the reader is referred to [37].

## III. Computational framework

The proposed computational framework consists of three principal stages: (i) data preparation and windowed indicator construction, (ii) dynamic risk assessment via a joint optimization strategy, and (iii) model aggregation and interpretation. Although we used i-DREAMS signals (Mobileye ADAS, GPS, CAN bus, TSR, and driver metadata) to test the framework, any time-stamped kinematic data can be fed into it after basic synchronization. In Stage 1, time-series are aligned, categorical variables encoded, missing values handled, trips sorted by driver identity for group-aware splitting, and rolling observation windows with paired prediction horizons are built and continuously updated throughout each trip to derive time-varying indicators such as speed-weighted headway and harsh events. In Stage 2, indicators are converted into binary risk labels using an adaptive threshold for harsh events that fires only when the signal stays beyond the limit for multiple consecutive samples (i.e., not a single-sample spike), and, for headway, either an adaptive threshold or a train-only quantile rule; the threshold is adjusted over time within driver-specific splits and trip segments using empirical-Bernstein uncertainty and a regret-tracking check that fine-tunes when the harmonic mean (HM) of accuracy and F1 of each model falls below the best-so-far level, keeping the system responsive yet stable. In Stage 3, Random Forest (RF), XGBoost (XGB), and Deep Neural Network (DNN) models are trained on SHAP-selected features; hyperparameters are optimized, validation probabilities are calibrated (isotonic or Platt/sigmoid, chosen by lower validation log-loss), model outputs are fused using validation-derived weights, and a single operating threshold is then chosen on validation by maximizing the HM of accuracy and F1. To obtain unbiased, leakage-safe estimates, a nested, driver-aware, time-ordered cross-validation (CV) is used for Stage 3: an inner loop (tunes hyperparameters, selects the calibration method, and sets the operating threshold using train-only statistics) and an outer loop (refits and evaluates on held-out drivers). Fig. 1 provides a schematic representation of the framework; detailed components and mathematical specifications are presented in the subsequent sections.

### *A. Data splitting and feature selection*

To ensure fair evaluation, prevent information leakage, and preserve the longitudinal structure of driving behavior, the dataset of Belgian drivers was partitioned and processed in a group- and time-aware manner. Drivers were assigned wholly to a single partition so that no individual appeared in both training and evaluation data. Trips within each driver were kept in chronological order to maintain the temporal flow of behavior. The full dataset was first sorted by driver identifier, trip identifier, and timestamp, then split in a group-aware manner into a primary training set (70%) and a temporary hold-out set (30%), ensuring driver disjointness throughout. The hold-out set was subsequently divided, again with the same group-aware logic, into validation (15%) and test (15%) subsets. All subsets were re-sorted chronologically by trip to maintain temporal consistency, and the target variable was re-aligned after splitting to ensure correct indexing.

Feature selection was conducted on a representative subset of the training data. To reduce computational load while preserving group structure and class distribution, the subset was first downsampled (with optional trip-based grouping) and then balanced using the Synthetic Minority Over-sampling Technique (SMOTE) [38]. A LightGBM classifier [39] was trained on this balanced subset, and SHAP (SHapley Additive exPlanations) [40] values were computed to rank predictors for the dynamic target. While SHAP values have been mostly used for interpretation the outputs of machine-learned models, it has also been found to show promising results when using it as a feature selection tool [41]. Candidate features were restricted to a predefined pool of top-ranked variables, and only those with SHAP importance above a minimum threshold(0.001) were retained. This selection step was coupled to the effective decision threshold of the main

TABLE I
DESCRIPTION OF VARIABLES EXTRACTED FROM I-DREAMS DATASET FOR THE FRAMEWORK

| Variable | Description | Unit | Type |
|---|---|---|---|
| Time headway (discrete) | Defined in three levels: safe headway (1: ≥ 2.5 s), danger headway (2: 0.6 s < hw < 2.5 s), and avoidable crash headway (3: < 0.6 s) | Proportion per TW* | Target/Predictor |
| Overtaking | Mean number of illegal overtaking events within TW | - | Predictor |
| Speeding | Defined in three levels: 1: speeding less than 10% over the limit, 2: speeding between 10% and 15% over the limit, and 3: speeding more than 15% over the limit within TW. | Proportion per TW | Predictor |
| Harsh acceleration | Proportion of harsh acceleration events (> 0.31 g) | Proportion per TW | Predictor |
| Harsh braking | Proportion of harsh braking events (> 0.31 g) | Proportion per TW | Target/Predictor |
| Harsh cornering | Proportion of critical lateral deviation | Proportion per TW | Target/Predictor |
| TSR Level (ADAS) | ADAS-detected Traffic Sign Recognition level; road environment proxy | Categorical (ordinal) | Predictor |
| Time headway (ADAS) | Headway estimate from ADAS (alternative to map-based headway) | Seconds | Predictor |
| Wiper Activity | Proxy for environmental condition (e.g. rain, reduced visibility) | Binary | Predictor |
| Trip Duration (s) | Total duration of the trip in seconds; captures overall exposure time and potential fatigue effects | Seconds | Predictor |
| Trip Index | A sequential identifier for trips driven by a participant, capturing temporal effects such as learning, fatigue accumulation, or session-specific variability | Integer | Predictor |
| IBI | Inter Beat Interval | Mean per TW | Predictor |
| Speed | Actual speed of the car in km/h | Mean per TW | Predictor |
| Age | Age of the driver | Years | Predictor |
| Gender | Male/Female | - | Predictor |
| Education | The highest level of education | - | Predictor |
| Driving experience | - | Years | Predictor |
| Income | Drivers' income | Euros | Predictor |
| Dominant environment of driving | Consists of three categories of rural (roads with a max speed limit of 70/90 km/h), urban (roads with a max speed limit of 30/50 km/h) and motorway (roads with a maximum speed limit 120 km/h) | - | Predictor |
| Driver's attitudes and beliefs | about speed limit, drowsiness, distraction, illegal overtaking, safe distance to the leading vehicle and interaction with pedestrians on roads | 5-point Likert scale | Predictor |
| Driver's confidence | about their driving skills and crash avoidance abilities | 5-point Likert scale | Predictor |
| Driver's attitudes and perceived benefits | of risky driving behaviors | 5-point Likert scale | Predictor |
| Driver's attitudes and perceived benefits | of safe driving | 5-point Likert scale | Predictor |
| Crash involvement | If they were involved in a car crash in the past three years | Binary | Predictor |
| Traffic offence | If they committed a traffic offence in the past three years | Binary | Predictor |

**Time window*

pipeline: SHAP-based selection was re-triggered only when evidence of a material shift was observed—specifically, when (i) the absolute change in the effective threshold ($\tau_t^{\text{eff}}$; see (16)) exceeded 0.01, or (ii) feature drift, measured by the Wasserstein distance (Panaretos & Zemel, 2019; Vallender, 1974) between successive training distributions, exceeded 0.05. This conditional re-selection avoided unnecessary recomputation while remaining responsive to meaningful changes in data or model behavior.

### *B. Risk Prediction Models*

Two conventional machine learning models and one deep learning model were employed within the computational framework to predict risky driving behaviors, based on a set of defined risk indicators. These models include a random forest (RF), an XGBoost (XGB) and a deep neural network (DNN) which are all explained in Sec. I of the supplementary material.

### *C. Hyperparameter Tuning*

Hyperparameter tuning for the models is performed employing Optuna [44] which is an open-source, automated hyperparameter optimization framework designed for efficient and flexible hyperparameter tuning. This framework is based on sequential model-based optimization (SMBO) [45], particularly Bayesian optimization with a Tree-structured Parzen Estimator (TPE) surrogate model [46], [47].

In addition to hyperparameter optimization, specific measures are also taken to mitigate overfitting during model training. The DNN model is trained using the Correlation-Dependent Stopping Criterion (CDSC), a dynamic early stopping mechanism based on the correlation between training and validation loss trends [48]. The CDSC formulation is as

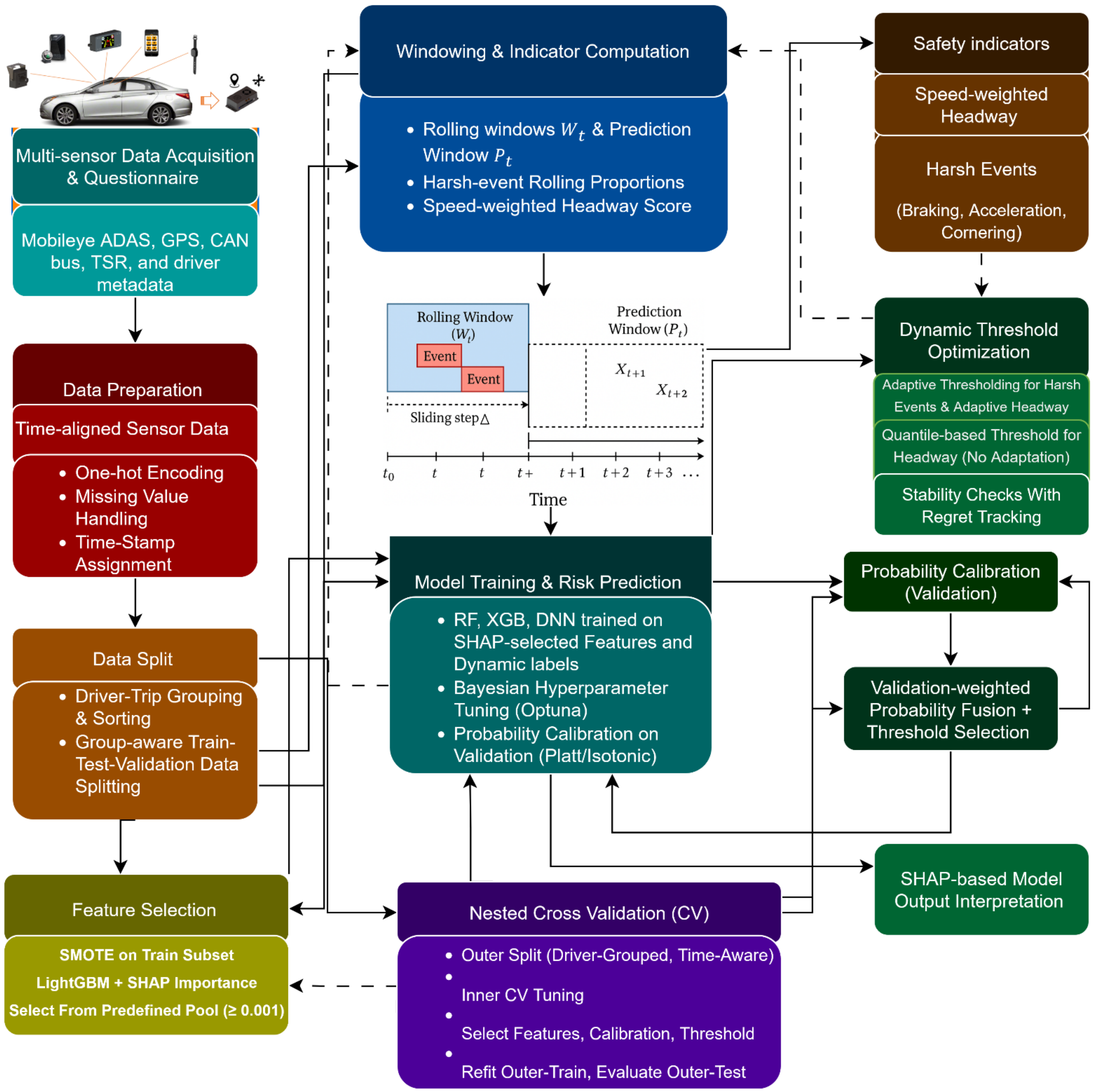

**Fig. 1.** Framework for risky-driving prediction from time-stamped data. RF/XGB/DNN outputs are probability-calibrated and fused with validation-derived weights; a single operating threshold is selected on validation; evaluation uses nested, subject-aware CV. ***Note:*** the data acquisition panel is illustrative and not part of the computational framework.

follows:

$$r = \frac{\sum_{j=n_i-\kappa}^{n_i}\left(e_{\mathrm{tr}}^{[j]}-\overline{e_{\mathrm{tr}}}\right)\left(e_{\mathrm{va}}^{[j]}-\overline{e_{\mathrm{va}}}\right)}{\sqrt{\sum_{j=n_i-\kappa}^{n_i}\left(e_{\mathrm{tr}}^{[j]}-\overline{e_{\mathrm{tr}}}\right)^2}\cdot\sqrt{\sum_{j=n_i-\kappa}^{n_i}\left(e_{\mathrm{va}}^{[j]}-\overline{e_{\mathrm{va}}}\right)^2}} \quad (1)$$

where $n_i$ is the current epoch index ( $i > \kappa, \kappa \geq 1$ ), $e_{\mathrm{tr}}^{[j]}$ and $e_{\mathrm{va}}^{[j]}$ denote the training and validation errors at epoch $j$, respectively and $(\overline{e_{\mathrm{tr}}})$ *and* $(\overline{e_{\mathrm{va}}})$ represent the mean training and validation errors within the rolling window from epoch $(n_i - \kappa)$ to $(n_i)$.

Similarly, to improve generalization and training efficiency in tree-based models, adaptive early stopping strategies are implemented, monitoring validation loss for XGB and dynamically tracking Out-Of-Bag (OOB) error in RF to guide model growth.

### *D. Risky driving classification within a rolling time window*

Risky-driving classification should be time-sensitive, with recent patterns and evolving trends captured. Past behavior is used to support real-time inference, whereas future behavior, although unavailable online, is used in offline analyses and for threshold calibration. Accordingly, a rolling window is used to continuously aggregate critical events, and two indicators are employed to classify risky episodes as they occur. Specifically, at any given time step $t \in \mathbb{Z}$, the data within a rolling window $W_t$ of duration $T$ seconds is utilised to capture the temporal evolution of driving behaviours. This window is defined in terms of a discrete length parameter $\omega$ which specifies the number of past time steps (in seconds) included in each observation window. Formally:

$$\mathrm{W_t} = \{\mathrm{X_{t-\omega+1}}, \dots, \mathrm{X_t}\} \tag{2}$$

As time progresses, the window slides forward by a fixed step size $\Delta$ , forming the next observation window:

$$\mathrm{W_{t+\Delta}} = \{\mathrm{X_{t-\omega+1+\Delta}}, \dots, \mathrm{X_{t+\Delta}}\} \tag{3}$$

In parallel, a prediction window of length $P$ seconds is defined to capture short-term future dynamics:

$$P_t = \{X_{t+1}, X_{t+2}, \dots, X_{t+P}\} \tag{4}$$

**Harsh events.** Let $\mathcal{X} = \{h_a, h_b, h_c\}$ denote harsh acceleration, braking and cornering with $x \in \mathcal{X}$. Event proportions are defined over the observation and prediction windows:

$$\mathrm{ev}_x(W_t) = \frac{1}{\omega}\sum_{i=t-\omega+1}^{t} \mathbb{1}_x(i), \tag{5a}$$

$$\mathrm{ev}_x(P_t) = \frac{1}{P}\sum_{i=t+1}^{t+P} \mathbb{1}_x(i) \tag{5b}$$

where $\mathbb{1}_x(i) \in \{0,1\}$ indicates whether event $x$ occurred at time $i$. A binary indicator is set to 1 if the event proportion exceeds the effective threshold in either window:

$$b_x(t) = 1\{ \, if \; ev_x(W_t) > \tau_t^{\mathrm{eff}} \;\vee\; ev_x(P_t) > \tau_t^{\mathrm{eff}}\} \tag{6}$$

here, $\tau_t^{\mathrm{eff}} \in [0,1]$ is the threshold that determines whether these proportions translate into risky driving (See also (16)). The harsh event indicator is:

$$R_{harsh}(t) = b_{ha}(t) \vee b_{hb}(t) \vee b_{hc}(t) \tag{7}$$

**Speed-weighted headway.** Critical headways represent sustained risk stemming from insufficient time gaps between vehicles and are especially hazardous at higher speeds, where reaction time and braking distance are constrained. To reflect this, we propose a speed-weighted headway scoring model as in the following:

Let $\mathrm{hdw}_k(t) \in \{0,1\}$ denote the binary flag for headway level $k \in \{1,2,3\}$, with $k = 3$ being most critical. The rolling critical headway proportion for each level is:

$$\mathrm{hdw}_k^{\mathrm{prop}}(t) = \frac{1}{\omega}\sum_{i=t-\omega+1}^{t} \mathrm{hdw}_k(i) \tag{8}$$

Let $v_t$ denote the mean speed within $W_t$. Define train-only empirical speed bounds using training-set percentiles to avoid leakage:

$$s_{\mathrm{low}} = F_{v,train}^{-1}(p_{low}), \quad s_{\mathrm{high}} = F_{v,train}^{-1}(p_{high}) \tag{9}$$

where $F_{v,train}^{-1}(\cdot)$ denotes the empirical quantile function of $\{v_t\}$ on the training set. For each level $k$, assign a tunable weight $\alpha_k \in \mathbb{R}^+$ such that $\alpha_1 < \alpha_2 < \alpha_3$ . Define the piecewise speed sensitivity function as:

$$w_k(v_t) \begin{cases} 0 & v_t \le s_{\mathrm{low}} \\ \alpha_k \cdot \frac{v_t - s_{\mathrm{low}}}{\max(s_{\mathrm{high}} - s_{\mathrm{low}},\ \varepsilon_s)} & s_{\mathrm{low}} < v_t < s_{\mathrm{high}} \\ \alpha_k & v_t \ge s_{\mathrm{high}} \end{cases} \tag{10}$$

where $\varepsilon_s > 0$ is a small stabilization constant and $p_{low}$ and $p_{\mathrm{high}}$ are percentile levels used to set $s_{\mathrm{low}}$ and $s_{\mathrm{high}}$, respectively.

The speed-weighted headway risk is:

$$\mathrm{RiskScore_{hdw}}(t) = \sum_{k=1}^{3} w_k(v_t) \cdot \mathrm{hdw}_k^{\mathrm{prop}}(t) \tag{11}$$

Two thresholding modes are considered:

**Quantile mode.** Label by comparing the risk score to a train-only quantile threshold $\sigma$:

$$R_{\mathrm{hdw}}(t) = 1 \; if \, \mathrm{RiskScore_{hdw}}(t) > \sigma$$

where:

$$\sigma = Q_\theta(\{\mathrm{RiskScore_{hdw}}(u) : u \in \mathcal{L}_{Train}\}), \theta \in (0,1) \tag{12a}$$

**Adaptive mode.** Normalize risk to [0,1], apply persistence and an adaptive threshold $\tau$:

$$r(t) = \frac{\mathrm{RiskScore_{hdw}}(t)}{\alpha_1 + \alpha_2 + \alpha_3} \in [0,1], \tag{12b}$$

$$r_{\mathrm{future}}(t) = \frac{1}{|P_t|}\sum_{j \in P_t} r(j) \tag{12c}$$

$$R_{\mathrm{hdw}}(t) = 1\{r(t) > \tau_t^{\mathrm{eff}} \vee r_{\mathrm{future}}(t) > \tau_t^{\mathrm{eff}}\} \tag{12d}$$

The unified indicator is:

$$R(t) = R_{\mathrm{harsh}}(t) \vee R_{\mathrm{hdw}}(t) \tag{13}$$

In this study, harsh and headway indicators are evaluated in separate pipelines (not fused into a single model).

**Adaptive thresholding with uncertainty control.** To calibrate risk decisions over time, an adaptive thresholding mechanism is introduced for label paths that use $\tau$ (both harsh events and adaptive speed-weighted headway), combining window statistics, an empirical-Bernstein margin, and regret-based updates. The components are defined below.

For any bounded signal $x(i) \in [0,1]$, define the mean and variance over the observation window as:

$$\overline{x}_W(t) = \frac{1}{\omega}\sum_{i \in W_t} x(i), \tag{14a}$$

$$\widehat{\sigma_W^2}(t) = \frac{1}{\omega}\sum_{i \in W_t}\left(x(i) - \overline{x}_W(t)\right)^2 \tag{14b}$$

To account for finite-sample and variance-dependent uncertainty in the windowed estimates, the base threshold is augmented by an empirical-Bernstein margin; with confidence parameter $\delta \in (0,1)$ and per window effective sample size

$n = |W_t| = \omega$ (i.e., the count within the rolling window), the margin is:

$$\rho_t = \sqrt{\frac{2\,\widehat{\sigma_W^2}(t)\,\ln(3/\delta)}{n}} + \frac{3\,\ln(3/\delta)}{n-1} \tag{15}$$

where $x(i) \in [0,1]$, and $\widehat{\sigma_W^2}$ denotes the biased (population) variance of $x$ over $W_t$. The effective threshold is then projected onto $[\tau_{min}, \tau_{max}] \subseteq [0,1]$ to stabilize the false-positive trade-off:

$$\tau_t^{eff} = min\{\tau_{max}, max\{\tau_{min}, \tau_t + \rho_t\}\} \tag{16}$$

Let $r_j \geq 0$ denote the regret logged at iteration $j$. The lookback horizon used for smoothing depends on whether the current label path uses $\tau_t$:

$$L = \begin{cases} \omega + \mathrm{P} & Harsh/Speed - weighted\ headway\ (adaptive)\ (uses\ \tau_t) \\ \omega, & Speed - weighted\ headway\ (quantile)\ (does\ not\ use\ \tau_t) \end{cases} \tag{17}$$

Let $\mathrm{HM}(a,f) = \frac{2af}{a+f}$ be the harmonic mean of accuracy and F1. Define non-negative regret:

$$r_j = \max 0, \mathrm{HM}^{\star} - \mathrm{HM_j} \in [0,1] \tag{17a}$$

where $\mathrm{HM}^{\star}$ is a reference (e.g., best-so-far). Updates use the mode-aware lookback $L$.

To stabilize threshold updates, recent optimization feedback is summarized using a rolling average regret and its dispersion over the mode-aware lookback $L$ (defined in (17)):

$$\mathrm{RoR_t} = \frac{1}{L}\sum_{j=t-L+1}^{t} r_j \tag{18}$$

$$\mathrm{RV_t} = \sqrt{\frac{1}{L}\sum_{j=t-L+1}^{t}\left(r_j - \mathrm{RoR_t}\right)^2} \tag{19}$$

where $\mathrm{RoR_t}$ denotes the rolling mean of recent regret and $\mathrm{RV_t}$ is its dispersion (population standard deviation) over the same horizon. A small $\mathrm{RV_t}$ indicates stabilization and gates further $\tau$-updates, while $\mathrm{RoR_t}$ and $\mathrm{RV_t}$ together inform the adaptive step size when updates proceed.

To prevent oscillatory updates, adaptation is suspended once short-term regret variability falls below a small tolerance $\varepsilon_r$; in that case the threshold is kept fixed:

$$\mathrm{RV_t} < \varepsilon_r \Rightarrow \tau_{t+1} = \tau_t \tag{20}$$

After computing the mode-aware regret statistics, the update step size $\eta$, which governs threshold movement is adapted according to short-term variability: it is cautiously increased when regret variation is high (to react faster) and decreased otherwise (to avoid jitter).

$$\eta_{t+1} = \begin{cases} \min\{\eta_{max}, \eta_t + \Delta_{\uparrow}, \mathrm{RV_t} > \beta \\ \max\{\eta_{min}, \eta_t - \Delta_{\downarrow}, \mathrm{RV_t} < \beta \end{cases} \tag{21}$$

where $\beta > 0$ is a small variability tolerance and $\Delta_{\uparrow}$, $\Delta_{\downarrow}$are fixed increments.

Given the refreshed step size $\eta_{t+1}$, the classification threshold $\tau$ is adjusted using instantaneous regret $r_t$ and a horizon factor $(1 + \frac{P}{\omega})$: positive regret moves $\tau$ downward (greater sensitivity), otherwise it nudges $\tau$ upward in proportion to the current performance gap; the result is clipped to $[\tau_{min}, \tau_{max}]$.

$$\tau_{t+1} = \begin{cases} \tau_t - \eta_{t+1} r_t (1 + \frac{P}{\omega}),\ r_t > \varepsilon \\ \tau_t + \eta_{t+1} r_t (1 - h_t^{*}),\ r_t < \varepsilon \end{cases} \tag{22}$$

where $\varepsilon > 0$ prevents spurious flips when regret is numerically near zero and $h_t^{*} \in [0,1]$ denotes the current harmonic-mean performance (higher is better).

A new target variable $(y)$ is materialised only when the threshold has moved sufficiently or when covariate shift is detected. Feature drift is monitored via the 1-Wasserstein distance between successive empirical feature distributions over the selected set:

$$|\tau_t - \tau_{t-1}| > \Delta_\tau \ \vee\ \max_f W_1\left(\widehat{F_{t-1}^f}, \widehat{F_t^f}\right) > \zeta \tag{23}$$

where $\Delta_\tau > 0$ is a minimal effective threshold change, $W_1(\cdot,\cdot)$ is the 1-Wasserstein distance, $\widehat{F_t^f}$ is the empirical distribution of feature $f$ at step $t$ and $\zeta > 0$ is a drift tolerance. When triggered, the target is recomputed and the SHAP-based feature set is optionally refreshed.

*E. Ensemble threshold optimization via calibrated probability fusion*

After model-wise training, the ensemble operates on probabilities rather than early binarized labels. The procedure comprises four conceptual stages: (i) tuning each model with a proper scoring rule, (ii) calibrating its probabilities on held-out data, (iii) fusing calibrated probabilities with validation-derived weights, and (iv) selecting a single operating threshold that maximizes a target metric. This is performed separately for each safety indicator (harsh, headway).

For model $f_i(\cdot;\theta_i)$, let $p_i(x;\theta_i) \in [0,1]$ denote the predicted probability for input $x$. Hyperparameters $\theta_i$ are chosen to minimize a scoring rule on a validation set $V$ (e.g., negative log-likelihood), without fixing any classification threshold:

$$\theta_i^{\star} = \arg\min_{\theta_i} \mathcal{S}_{\mathrm{V}}\left(\widehat{p_i}(\cdot;\theta_i)\right), \tag{24a}$$

where:

$$\mathcal{S}_{\mathrm{NLL}}(V) = -\frac{1}{|V|}\sum_{(x,y)\in V}[y \log \widehat{p_i}(x) + (1-y)\log(1 - \widehat{p_i}(x))] \tag{24b}$$

and $V = \{(x,y)\}$ is the held-out validation set, $y \in \{0,1\}$ is the label, $\widehat{p_i}(\cdot;\theta_i)$ is the model's predicted probability, $|V|$ is the number of validation samples, and $\mathcal{S}_{\mathrm{V}}(\cdot)$ is a proper scoring rule.

To improve threshold transferability and comparability across models, each tuned model is calibrated on a held-out split via a monotone map $\phi_i: [0,1] \to [0,1]$ (Platt scaling/isotonic regression):

$$\widetilde{p_i}(x) = \phi_i(\widehat{p_i}(\cdot;\theta_i^{*})) \tag{25}$$

Calibrated probabilities are then combined using non-negative validation-derived weights $\lambda_i$ (e.g., proportional to a calibration-set performance score), yielding the ensemble probability:

$$\bar{p}(x) = \left(\frac{\sum_{i=1}^{m} \lambda_i \cdot \widehat{p_i}(x)}{\sum_{i=1}^{m} \lambda_i}\right) \tag{26}$$

A single decision threshold $\Theta \in (0,1)$ is selected on validation data to maximize the HM of accuracy and F1 over a candidate grid $G \subset (0,1)$:

$$\Theta^* = \arg\max_{\Theta \in G} \text{HM}\left(\text{Acc}(\Theta), \text{F1}(\Theta)\right), \tag{27a}$$

$$C(x; \Theta^*) = 1\{\bar{p}(x) \geq \Theta^*\} \tag{27b}$$

where all variables are introduced previously.

### *F. Nested, subject-aware validation*

We use nested CV with subject-disjoint outer folds ($K$), so each outer test split contains unseen drivers, and time-ordered inner folds ($J$) built within each outer-train partition. Validation blocks are aligned to complete trips ensuring that no trip is split across training and validation folds. A purge/embargo of length $E = |W| + |P| - 1$ is applied around each validation block to prevent look-ahead and temporal spillover between the observation. Binary labels are recomputed per fold from train-only statistics (Sec. III-D), ensuring no validation information leaks into label construction. Model hyperparameters are tuned to minimize the mean validation negative log-likelihood (NLL). Inner probabilities are calibrated (isotonic or Platt, chosen by lower validation log-loss), and a single operating threshold $\Theta^*$ is chosen on validation by maximizing the HM of Accuracy and F1, then held fixed. Final models are refit on outer-train; calibration is re-estimated on outer-train using driver-aware out-of-fold predictions, then fixed and applied to outer-test. Metrics are reported at the subject level. Full details appear in Sec. II of the supplementary material.

## III. RESULTS

### *A. Implementation details*

The computational framework was implemented with a rolling window $T = 5\ s$, sliding step $\Delta = 1\ s$, and prediction window $P = 2\ s$. The observation window was set to 5 s to provide context beyond the typical 1–2.5 s perception–reaction interval used in design and to cover short-horizon driver adaptation (often ~4–6 s under changing demands) [49]. Here, we opted to vary the risk classification threshold ($\tau_t$) while keeping the window size fixed, rather than making the window size itself adaptive. Varying window sizes introduces additional variability and can complicate the temporal alignment of driving events across trips. This approach reduces the dimensionality of the optimization space, limits overfitting to variable-length patterns, and provides more stable convergence. In essence, the fixed window provides a consistent perceptual frame, while the threshold ($\tau_t^{\text{eff}}$) evolves as a flexible decision boundary. Nevertheless, all of these parameters could be exposed as tunable components within the broader optimization framework if desired.

$\tau_t$ was initialized at $t = 0$ to 0.5 to derive provisional binary labels per window. Optimization comprised 25 Optuna trials per indicator (harsh events; speed-weighted headway), jointly tuning model hyperparameters and adapting $\tau_t$ (via the uncertainty-aware, regret-guided updates). For speed-weighted headway we implemented two labelling paths—(i) a train-only quantile rule and (ii) the adaptive $\tau_t$ rule—but, to preserve comparability with the harsh-event pipeline (which also uses adaptive $\tau_t$), we reported headline results for the adaptive headway variant; quantile-mode analyses were retained for robustness and CV checks. Per-model training used: (i) DNN with ReduceLROnPlateau and CDSC, restoring the best validation-loss checkpoint; (ii) RF with warm-start growth and early stopping on OOB score; and (iii) XGB with native early stopping on validation log-loss; all with Optuna pruning. After training, per-model thresholds ($\Theta_i^*$, chosen to maximize the HM of accuracy and F1 on validation) were used only for per-model reporting and did not influence the adaptive label threshold $\tau_t$ (or its effective form $\tau_t^{\text{eff}}$). Separately, for the ensemble we calibrated each model's validation scores (isotonic regression when sample size/score variability permitted; otherwise Platt scaling), formed validation-derived non-negative fusion weights (inverse validation log-loss), and fused calibrated probabilities. A single ensemble operating threshold $\Theta^*$ was then selected on validation by maximizing the same HM criterion and held fixed for test evaluation (Sec. III-E).

For nested CV, the outer loop used five folds by driver, so each test fold contained drivers unseen in training. Within each outer training set, model tuning and threshold selection used two time-ordered validation splits. To prevent look-ahead from the rolling (5 s) and prediction (2 s) windows, we applied a 6-second embargo around each inner validation block (5 s + 2 s − 1 s). Validation blocks respected trip boundaries. All train-only statistics (e.g., percentiles for speed bounds), the dynamic labels, and the single operating threshold selected on validation were fixed before evaluating on the outer test fold. Final results were reported as the mean ± SD across the five outer folds.

Experiments were executed on the DelftBlue Supercomputer (2× NVIDIA A100 GPUs, 50 CPU cores, CUDA 12.2). This environment accelerated hyperparameter search and ensemble calibration; the pipeline remains executable on modest hardware at higher runtimes. All the evaluation metrics are defined in Sec. III of the supplementary material.

### *B. Model Calibration, threshold adaptation, and convergence properties*

The results of hyperparameter tuning, including the hyperparameters used, their search spaces, and the best values identified in the Optuna trials, can be found in Sec. IV of the supplementary material. Fig. 2 presents the effect of dynamically selected threshold values ($\tau_t$) on HM across trials for each model, separately for harsh events (top) and critical headways (bottom). For harsh event detection, all three models achieved optimal HM values at lower thresholds ($\tau_t \approx$ 0.50), highlighting their higher discriminative power when the system is more sensitive to positive instances. However, DNN showed a notable performance decline as $\tau_t$ increased beyond 0.65, suggesting a reduced capacity to identify true positives under stricter classification regimes. Conversely, XGB maintained a consistently high HM across the threshold spectrum, suggesting that it is less sensitive to marginal threshold shifts. A similar but more pronounced pattern emerged in the headway domain, where the DNN achieved peak HM at lower thresholds ($\tau_t$< 0.50), but its performance was more variable across $\tau_t$ values, indicating sensitivity to calibration in this safety indicator. Notably, in the headway detection task, both RF and XGB models showed optimal performance only within a narrower, higher range of $\tau_t$ values (≥ 0.45). This indicates that, under the strong class imbalance of the dataset, the optimization process favored more conservative decision boundaries to reduce false positives, in contrast to the broader $\tau_t$ sweep observed in the harsh events domain. Moreover, RF's curve showed increased volatility in the headway context, particularly beyond $\tau_t = 0.50$, consistent with a threshold-dependent detection mechanism. This volatility was quantified as a standard deviation of 0.037 in HM, a maximum swing of 0.147 between consecutive thresholds, and a mean swing of 0.025. Meanwhile, XGB once again demonstrated a flatter and more stable HM profile, indicating its suitability for integration into dynamic thresholding architectures where $\tau_t$ adapts in response to context-specific regret signals (i.e., regret values computed within the rolling and prediction windows for the headway detection context) as well as shifts in recent event distributions. Collectively, these findings confirm that lower thresholds enhance model responsiveness to rare but critical behaviours, particularly for RF and DNN in the harsh events domain and for DNN in the headway domain, whereas XGB's stability across $\tau_t$ suggests it can act as an anchor model in ensemble configurations. See also Sec. V and Fig. SI of the supplementary material for the models' regret across optimization trials for both safety indicators.

### *C. Risky driving prediction*

Fig. 3 and Table II compare the performance of the three models across key evaluation metrics, separately for harsh events and critical headways. In the harsh events domain, the DNN model achieved the highest F1 score (0.916), recall (0.978), and harmonic mean (HM = 0.884), reflecting strong sensitivity, albeit partly due to its reliance on threshold tuning. The XGB model showed more balanced performance, with high precision (0.898), a relatively strong Matthews Correlation Coefficient MCC (0.398), and a robust Area Under the Precision–Recall Curve (AUC-PR) (0.888). Although the RF model showed comparatively lower overall performance (HM = 0.846), it achieved the highest AUC-PR (0.897), suggesting effective detection of positive instances under class imbalance despite its conservative predictions. In the critical headway domain, model performance converged more closely in terms of F1 scores (≈0.86) and HM (≈0.90), though subtle differences persist. XGB and RF showed superior recall (0.924 and 0.924 respectively) and MCC (~0.83), suggesting more reliable detection of rare events under calibrated thresholds.

DNN maintained competitive performance, particularly in precision (0.927), but showed lower recall and slightly reduced MCC, supporting earlier findings of volatility under threshold shifts. Overall, DNN achieved the highest HM across both harsh events and critical headways. The speed-weighted headway indicator emerged as the stronger predictor of risky driving behavior, with MCC approximately double that for harsh events. This likely reflects its more stable threshold adaptation and a better balance of true positives to true negatives under severe class imbalance.

Consistent with these patterns, the selected operating thresholds were $\Theta^* = 0.5$ for harsh events (yielding HM = 0.883) and $\Theta^* \approx 0.46$ for speed-weighted headways (yielding HM = 0.901). With a 5-s observation window, 0.5 for harsh events roughly corresponds to half the window meeting the event condition; for speed- weighted headway, $\Theta^*$ applies to the normalized risk score rather than a duration in seconds.

See Sec. VI of the supplementary material for SHAP-based interpretation of model outputs.

### *D. Nested CV results*

Table III presents the aggregate nested CV results, averaged across outer folds with isotonic calibration and frozen validation thresholds ($\Theta^*$). As in the main pipeline (Table II), RF and XGB performed closely in the harsh events domain, with HMs around 0.87 and F1 scores just above 0.90. RF again achieved the highest AUC-PR (0.942), while XGB produced nearly identical F1 and HM values (0.907 and 0.873, respectively). DNN's performance slightly improved over both tree-based models, achieving the highest F1 (0.939) and HM (0.909), confirming its advantage when thresholds are optimized across folds. Compared with Table II, these results show modest gains in AUC-PR and HM, reflecting the benefit of nested threshold calibration and driver-disjoint validation in reducing overfitting. Performance improved even further in the speed-weighted headway domain: all three models reached HM values near 0.93, with MCC values roughly double those seen for harsh events (≈ 0.82–0.84) and very high accuracies (≈ 0.976–0.977). This confirms that headway-based indicators yield more stable classification under severe class imbalance and that the proposed framework generalizes well beyond a single train/validation/test split.

## IV. Discussion

In this study, we developed a context-specific computational framework for predicting risky driving from naturalistic data, integrating adaptive thresholds and regret-based optimization. The above findings demonstrate two main advantages of our proposed framework: Firstly, by moving beyond static risk classification, the framework enables early (real-time) detection of risky driving events through adaptive threshold applied to driving behavior indicators within a continuously evolving temporal window, rather than relying solely on post-event analysis.

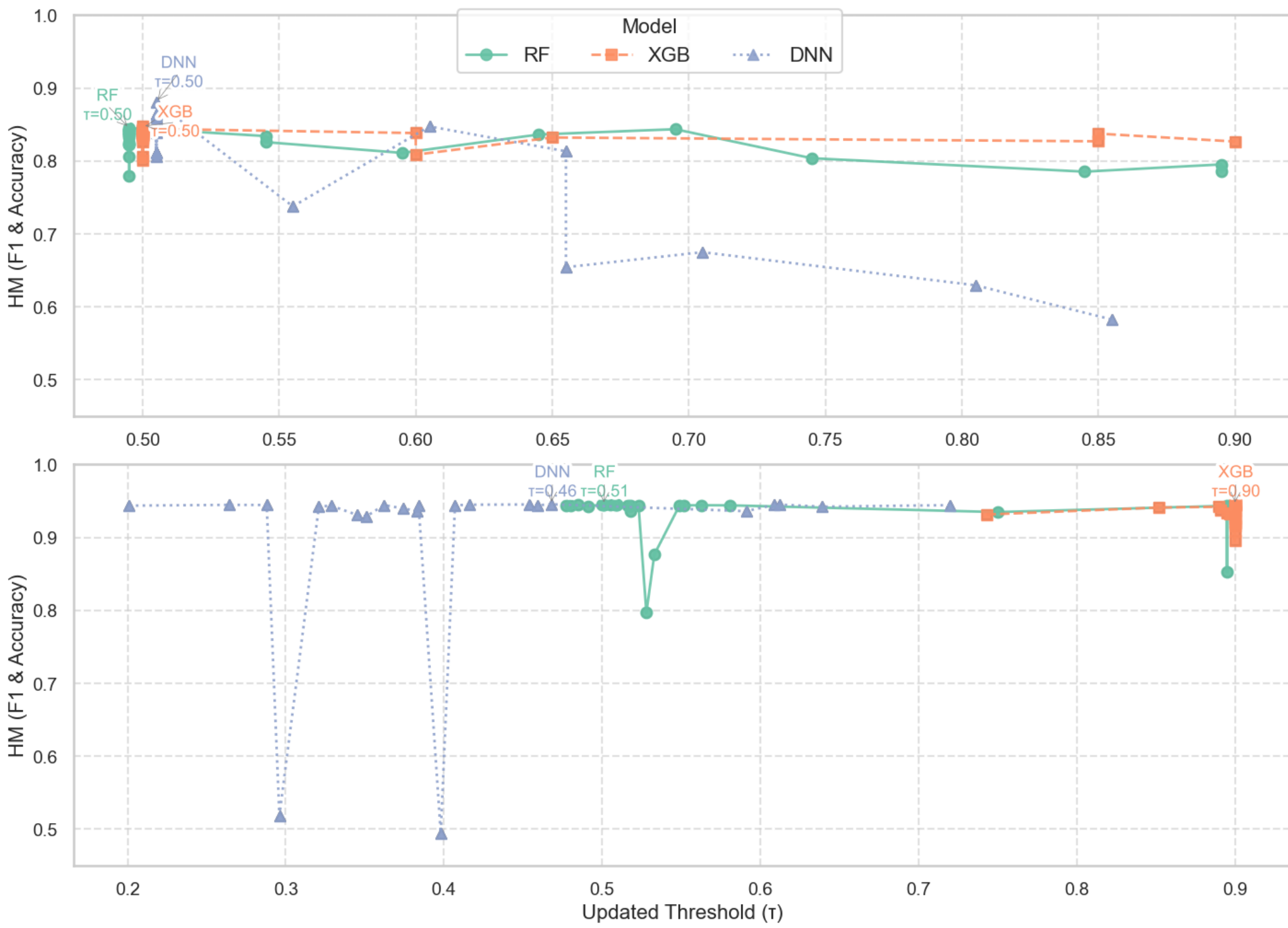

**Fig. 2.** Harmonic mean of F1 and accuracy as a function of the selected thresholds for harsh events (top) and headways (bottom).

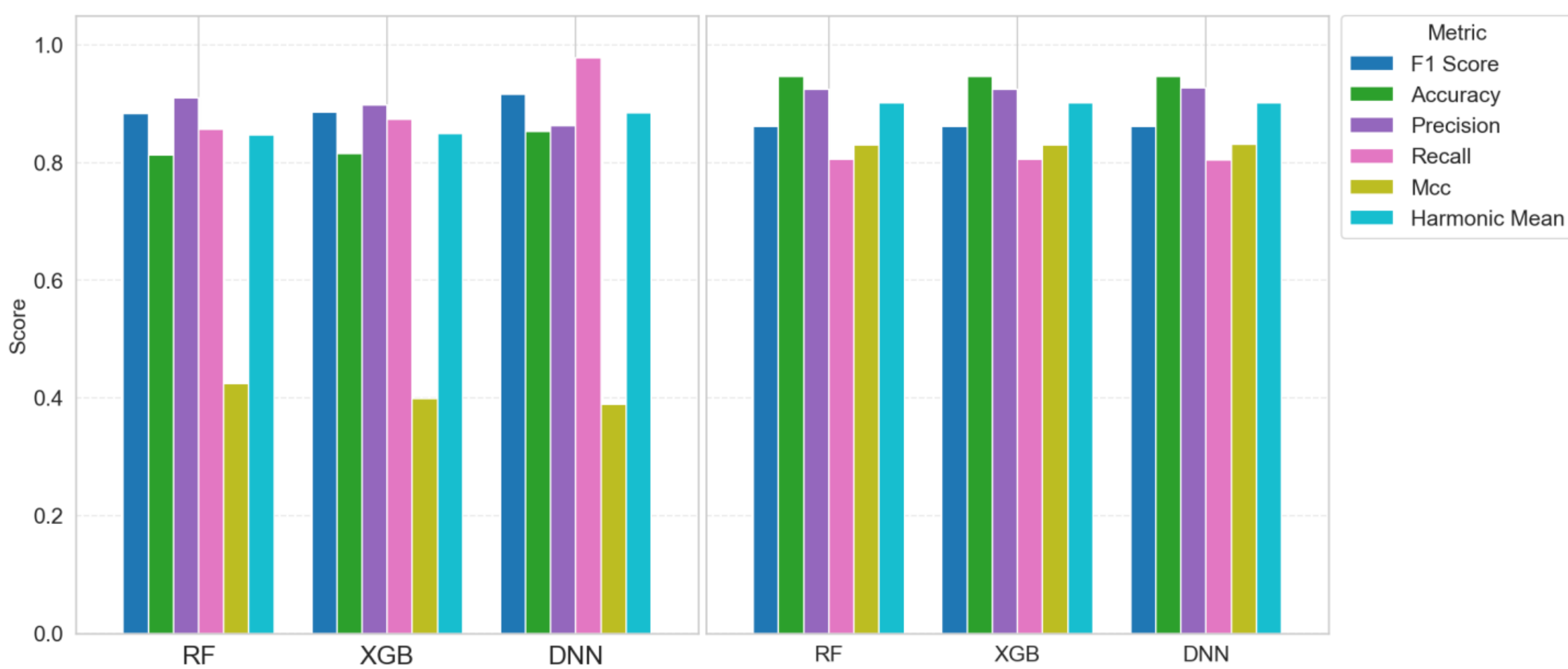

**Fig. 3.** Model comparison across harsh events (left) and critical headways (right).

TABLE II
PERFORMANCE METRICS FOR RF, XGB, AND DNN MODELS FOR HARSH EVENTS AND CRITICAL HEADWAYS

| Model | $\Theta_i^*$ | F1 Score | Accuracy | Precision | Recall | MCC | AUC-PR | HM |
|---|---|---|---|---|---|---|---|---|
| **Harsh events** | | | | | | | | |
| RF | 0.50 | 0.882 | 0.813 | **0.909** | 0.857 | **0.425** | **0.897** | 0.846 |
| XGB | 0.50 | 0.886 | 0.815 | 0.898 | 0.874 | 0.398 | 0.888 | 0.849 |
| DNN | 0.50 | **0.916** | **0.853** | 0.862 | **0.977** | 0.389 | 0.861 | **0.883** |
| **Speed-weighted headway** | | | | | | | | |
| RF | 0.50 | 0.860 | 0.945 | 0.924 | **0.805** | 0.820 | 0.785 | 0.901 |
| XGB | 0.90 | 0.860 | 0.945 | 0.924 | **0.805** | 0.830 | 0.774 | 0.901 |
| DNN | 0.46 | **0.861** | **0.946** | **0.927** | 0.804 | **0.831** | **0.786** | 0.901 |

TABLE III
NESTED CV PERFORMANCE OF ALL MODELS FOR HARSH EVENTS AND SPEED-WEIGHTED HEADWAYS

| Indicator | Model | AUC-PR | HM | F1 Score | Accuracy | MCC |
|---|---|---|---|---|---|---|
| Harsh events | RF | **0.942** | 0.872 | 0.906 | 0.842 | **0.427** |
| | XGB | 0.937 | 0.873 | 0.907 | 0.841 | 0.398 |
| | DNN | 0.904 | **0.909** | **0.939** | **0.882** | 0.391 |
| Headway | RF | 0.824 | 0.928 | 0.881 | **0.977** | 0.824 |
| | XGB | 0.816 | 0.928 | 0.881 | 0.976 | 0.830 |
| | DNN | **0.825** | **0.929** | **0.883** | 0.976 | **0.835** |

This dynamic risk assessment approach can support the generation of adaptive intervention triggers for ADAS and early risk warnings for human drivers which can improve system responsiveness and occupant safety [50]. Secondly, the modular architecture of the framework allows the individual components to be tuned, extended, or replaced with minimal impact on the overall system. This modular design ensures that modifications to one part of the framework can be implemented without necessitating structural changes to other components. The design flexibility enhances maintainability, supports context-specific adaptation, and offers compatibility across diverse risk modeling scenarios. Therefore, although the current article focuses on two traffic safety risk indicators, the framework is easily modifiable and scalable to incorporate other types of kinematic-dependent metrics such as TTC, acceleration/deceleration rates, PET, and various behavioral indicators.

The stronger performance of DNN in recall-based metrics for both harsh events and speed-weighted headway reflected recall-oriented threshold calibration, but it also showed greater trial-level volatility, indicating a trade-off between sensitivity and stability. By contrast, the overall performance of XGB highlights its suitability as a structurally robust and adaptable classifier for dynamic risk detection systems. Its relative insensitivity to moderate threshold shifts demonstrates algorithmic resilience, a property particularly valuable in real-world deployments where prediction thresholds evolve over time. This stability supports its integration into adaptive safety systems that require both interpretability and consistent performance under changing behavioral inputs [51]. In dynamic driving environments, behavioral data are inherently non-stationary due to contextual shifts in road type (e.g., transitioning from a motorway to a rural road), traffic flow (e.g., encountering a shockwave), weather conditions (e.g., changing from clear to rainy), and driver state (e.g., fatigue or distraction). Models deployed in these contexts must balance fine-grained behavioral responsiveness with long-term decision stability to avoid overfitting to transient noise [52]. Nonetheless, XGB showed slightly higher regret values in the headway domain compared with harsh events, suggesting that although the model handles outcome sparsity and class imbalance well, its optimization remains sensitive to distributional shifts and the specific relationships between features and outcomes. These findings highlight the importance of context-aware calibration strategies: leveraging XGB's stability in domains with consistent signal structure while complementing it with more agile models such as DNNs when heightened responsiveness is required, achieving the balance between sensitivity and robustness that prior research has identified as essential in dynamically evolving driving environments [53], [54].

Overall, each model showed distinct strengths aligned with the demands of the detection task: DNN achieved the highest recall-based performance for harsh events, indicating strong sensitivity to risky behaviors but at the cost of greater threshold-dependent variability; XGB maintained stable F1 and HM across thresholds and folds, making it a reliable backbone for adaptive ensemble integration, though its slightly higher regret in headway detection signals residual sensitivity to distributional shifts; and RF, despite being the most threshold-volatile in headway detection, delivered the highest precision and AUC-PR, making it effective for reducing false positives under severe class imbalance.

Our findings on the superiority of speed-weighted headway as a more stable and context-sensitive risk indicator than harsh driving events demonstrate a possible solution to a long-lasting conundrum in traffic safety: is headway a SSM? While foundational studies such as [55] have shown that short time headways significantly increase crash risk, later critiques, notably [16], highlighted headway's limitations, particularly its inability to account for lead-vehicle behavior dynamic closing speeds. By integrating speed directly into headway computation, our approach accounts for both spatial and kinematic dimensions of risk, especially in high-speed car-following contexts where brief gaps become more dangerous. This refinement is motivated by growing concerns over the limitations of traditional SSMs such as TTC, PET, and DRAC, particularly under car-following scenarios with minimal relative velocity. In such conditions, these classical indicators often lose discriminatory power, as TTC may approach infinity and DRAC may offer little predictive value when no braking is required [28]. Against this backdrop, speed-weighted headway offers a computationally efficient and conceptually intuitive alternative, especially for real-time applications where only ego-vehicle data are available. By directly integrating instantaneous velocity into the headway metric, it preserves sensitivity to elevated risk during high-speed short-gap following. However, it must be acknowledged that this approach does not account for lead-vehicle dynamics and, while tested here across both urban and highway conditions, it has not yet undergone widespread validation across broader traffic contexts, including variations in traffic density, driving culture, and roadway designs not represented in the present dataset.

This study has some limitations. The proposed framework was tested exclusively on a subset of the i-DREAMS dataset, Belgian drivers. Future research should evaluate the framework across additional country-specific subsets (e.g., Greek, German drivers) to assess its robustness to population heterogeneity, including behavioral, cultural, and contextual variability. This would help determine whether the dynamic thresholding mechanism adapts similarly across diverse conditions or if its optimization varies in response to different driving norms, infrastructure types, and operational environments. In this context, it would be particularly insightful to examine how dynamically estimated thresholds fluctuate relative to fixed baseline thresholds across heterogeneous populations. Additionally, the absence of commonly used SSMs, such as TTC and PET (which capture interactions with surrounding vehicles) limited our ability to benchmark the proposed harsh event and headway indicators against established safety proxies. Incorporating such interaction-based SSMs in future datasets would enable more comprehensive validation of our framework's predictive accuracy and behavioral sensitivity. Moreover, it is worth noting that while this study employed three widely used yet relatively conventional machine learning models (RF, XGB, and DNN), the choice was deliberate to demonstrate the robustness and generalizability of the proposed framework using interpretable and computationally efficient techniques. This ensured that the benefits of the framework such as dynamic threshold calibration, rolling-window adaptation, and regret-aware optimization were attributable to the system design rather than to model complexity. Although more advanced deep learning architectures, including Long Short-Term Memory networks (LSTMs), Convolutional Neural Network–Recurrent Neural Network hybrids (CNN-RNN hybrids), and Transformer models, have shown strong performance in learning temporal dependencies in behavioral data, the objective of the framework was not solely to maximize sequence modeling. Rather, it was to balance risk detection sensitivity with contextual adaptability and explainability, which are essential for real-time safety applications. Sequence-based models, while powerful for raw time-series analysis, often lack the transparency, adaptive thresholding, and computational efficiency necessary for operational deployment in dynamic driving contexts. Although recent Transformer-based models incorporate attention mechanisms that offer partial interpretability, they still fall short of providing the direct, threshold-based transparency and context-specific adaptability prioritized in our unified framework. As such, the proposed framework prioritizes interpretability and adaptive learning, offering a flexible foundation that can be extended and improved with hybrid architectures in future research.

## V. Conclusions

This study examined the potential of naturalistic driving data to support more nuanced identification of risky behaviors, moving beyond static thresholds and one-size-fits-all models. While sensor-rich data from real-world trips offer detailed insights into how individuals drive across diverse contexts, they also pose modeling challenges due to noise, behavioral heterogeneity, and temporal dependencies. To navigate this, we introduced a dynamic and context-aware framework that integrates empirical risk profiling with regret-based optimization to adapt risk thresholds over time and across drivers. Evaluated across two distinct behavioral safety indicators: harsh maneuvers and speed-weighted headways, the framework demonstrated robust performance under nested CV, balancing sensitivity to risky behaviors with protection against overfitting.

We evaluated RF, XGB, and DNN within the proposed framework and found all three to be competitive in predicting risky driving from harsh events and headway indicators. DNN offered the strongest sensitivity to elevated-risk conditions, RF was most effective at reducing false positives under severe class imbalance, and XGB provided the most stable performance across thresholds and folds. These complementary strengths highlight the value of combining models in adaptive ensemble systems for robust risk detection.

By incorporating interpretable machine-learning techniques (e.g., SHAP), the framework provides transparent insights into behavioral and contextual drivers of risk, supporting diagnostic feedback and model accountability. Its adaptability across time, driver profiles, and trip contexts suggests suitability for near-real-time deployment in personalized driver monitoring and intelligent vehicle applications, provided that deployment-specific latency and calibration requirements are met. Future work should validate the approach in broader geographical settings with more diverse driver populations and roadway environments to assess scalability and utility for next-generation road-safety interventions.

## Acknowledgments

The authors acknowledge the use of computational resources of the DelftBlue supercomputer, provided by Delft High Performance Computing Centre (https://www.tudelft.nl/dhpc). The authors would like to thank M. Azarmi for providing valuable feedback on the manuscript.

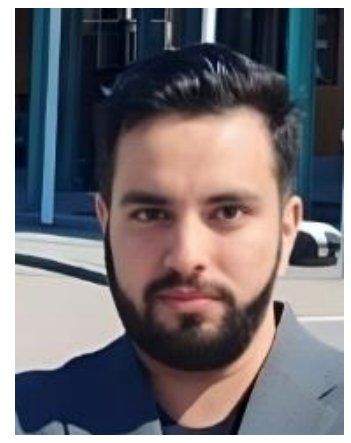

**Amir Hossein Kalantari** received the B.Sc. degree in Civil Engineering in 2015 and the M.Sc. degree in Transport Engineering in 2018. He obtained his Ph.D. in Transport Studies from the University of Leeds, U.K in 2023. He is currently a Postdoctoral Research Fellow at TU Delft, the Netherlands. His research interests include road user behavior modelling and driver monitoring, with a particular focus on human factors, vehicle automation, and the safety of vulnerable road users (VRUs). He has been a member of the American Society of Civil Engineers (ASCE) since 2015.

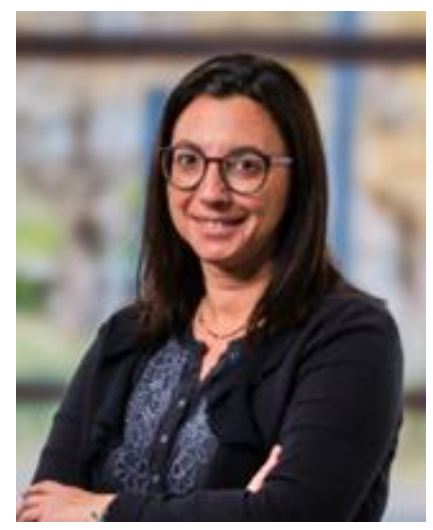

**Eleonora Papadimitriou** is Assistant Professor of Urban Mobility & Road Safety at the Department of Transportation Planning Engineering at the National Technical University of Athens (NTUA). From 2018 to 2024 she was Associate Professor of Transport Safety at Delft University of Technology (TU Delft). She has more than 20 years of experience in transportation engineering and road safety, with emphasis on active modes of transport and vulnerable road users, and on quantitative methods of risk and efficiency assessment.

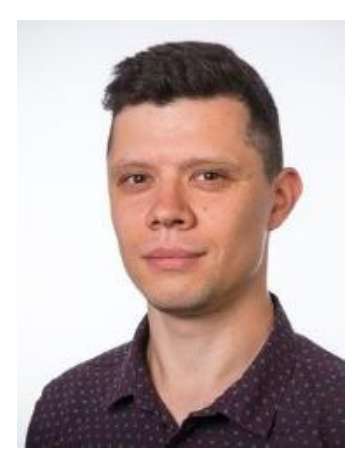

**Arkady Zgonnikov** is an assistant professor in the Department of Cognitive Robotics, TU Delft. He received his M.Sc. in applied mathematics from Saint-Petersburg State University, Russia, and Ph.D. in computer science and engineering from the University of Aizu, Japan. His work centers around modeling human behavior in traffic interactions and applying these insights to the development and evaluation of automated vehicles. He is also actively advancing the concept of meaningful human control over intelligent systems, both as a theoretical framework and as a practical principle for AI design and assessment, with a particular focus on automated vehicles.

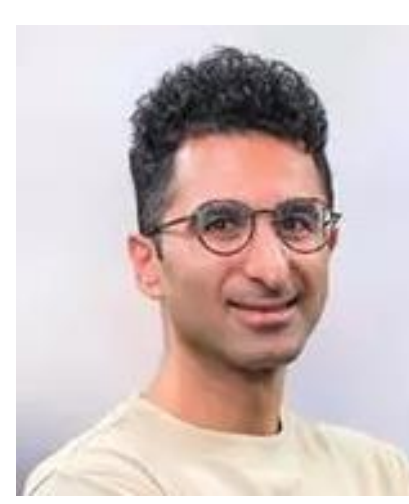

**Amir Pooyan Afghari** is an assistant professor of probabilistic methods in transport safety at TU Delft. His research interest is at the intersection of transport engineering, social science and data science: understanding human behavior and its impact on road safety using observational data, statistical and econometric methods and machine learning algorithms.

# Supplementary Material for "How Much Is Too Much? Adaptive, Context-Aware Risk Detection in Naturalistic Driving"

Amir Hossein Kalantari, Eleonora Papadimitriou, Arkady Zgonnikov, and Amir Pooyan Afghari

## I. Risk Prediction Models

### A. Random Forest (RF)

Random forests are based on bagging (bootstrap aggregating) in which multiple decision trees are trained on different subsets of the data, and their predictions are combined to improve performance and reduce overfitting. Each tree makes predictions by recursively splitting the data based on features, with splits chosen to maximize metrics such as Gini impurity or entropy. Once all trees are trained, the model uses majority voting (for classification) across all the trees' predictions [1].

### B. eXtreme gradient boosting (XGBoost)

XGBoost is an optimized implementation of the gradient boosting algorithm, offering improved computational efficiency and enhanced regularization capabilities to reduce overfitting, compared to the original formulation of gradient boosting. XGBoost includes L1 (lasso) and L2 (ridge) regularization terms in the loss function, which helps prevent overfitting by penalizing large weights. This model minimizes a custom loss function, typically log-loss for classification [2].

### C. Deep Neural Network (DNN)

A DNN is a type of artificial neural network (ANN) that consists of multiple hidden layers between the input and output layers, enabling it to capture complex non-linear patterns in the data. In this study, we implement a fully connected feed-forward architecture with between one and five hidden layers. Each layer applies a linear weighted transformation with He-uniform initialization, followed by a non-linear activation function. To enhance stability and generalization, batch normalization is applied after each dense layer, while dropout and L2 regularization are incorporated to mitigate overfitting. The network is trained end-to-end with binary cross-entropy loss and an optimizer selected from Adam, SGD (with momentum/Nesterov), or RMSprop. The final output layer employs a sigmoid activation to produce probability scores for binary classification, suitable for threshold-based risk detection.

## II. Nested Cross-Validation

We define the dataset as

$$D = \{(x_i, y_i, g_i, t_i)\}_{i=1}^{n},\ x_i \in \mathbb{R}^p,\ y_i \in \{0,1\} \tag{S1}$$

where $x_i$ are features and $y_i$ are labels from the main pipeline (Sec. III-D in the main manuscript) using nominal level $\tau_t$ and effective margin $\tau_i^{\text{eff}}$ (defined below). Here, $g_i$ is driver (group) ID and $t_i$ is time.

**Outer subject split.** For $k = 1, \dots, K$, form subject-disjoint pairs $(D_k^{\text{tr}}, D_k^{\text{te}})$ with

$$g(D_k^{\text{tr}}) \cap g(D_k^{\text{te}}) = \emptyset, \tag{S2}$$

thus the outer test measures out-of-subject generalization.

**Inner time-respecting CV.** Inside $D_k^{\mathrm{tr}}$, create J time-ordered folds

$$\left(D_{k,j}^{\mathrm{TR}},\, D_{k,j}^{\mathrm{VA}}\right), \qquad j = 1, \dots, J, \qquad D_{k,j}^{\mathrm{TR}} \cap D_{k,j}^{\mathrm{VA}} = \emptyset, \tag{S3}$$

and apply purge/embargo

$$E = |W| + |P| - 1 \tag{S4}$$

around each validation block (observation window |W|, prediction window |P| to remove look-ahead and temporal bleed-through). Validation blocks are aligned to complete trips.

**Diagnostics for leakage.** To prevent information leakage, we enforce the following conditions: (i) no identity overlap occurs at the outer level; (ii) validation timestamps always follow training timestamps inside each inner split; and (iii) no samples from the purge band appear on either side of a validation block.

**Dynamic Labels and Effective Threshold.** Dynamic labels are recomputed fold-locally from train-only statistics. The uncertainty-aware decision level uses an empirical-Bernstein radius $r_i$ (rolling, train-only):

$$\tau_i^{\mathrm{eff}} = \tau_t + r_i, \;\; \tau_t \in [\tau_{min}, \tau_{max}] \tag{S5}$$

The label rule $y_i\ (\tau_t,\ \tau_i^{\mathrm{eff}})$ covers harsh-event channels (adaptive, persistence-aware) and headway (either train-only quantile binarization or adaptive). All statistics are recomputed on $D_{k,j}^{\mathrm{tr}}$ and then applied to $\left(D_{k,j}^{\mathrm{tr}},\, D_{k,j}^{\mathrm{va}}\right)$.

Dynamic labeling can occasionally produce degenerate folds. In such cases, when an inner training fold reduces to a single class after dynamic labelling, we address this by (i) skipping SHAP and using a variance-based numeric subset; (ii) using model-native class weighting instead of resampling during tuning; (iii) still computing calibrated probabilities for threshold selection.

**Inner-Loop τ-Adaptation (Short-Horizon Regret).** Within each fold $(k, j)$, the effective threshold $\tau_t^{\mathrm{eff}}$ is adapted iteratively using a baseline model trained on $D_{k,j}^{\mathrm{tr}}$ and validated on $D_{k,j}^{\mathrm{va}}$. Candidate thresholds are projected within $[\tau_{min}, \tau_{max}]$ and adjusted proportionally to the horizon ratio $(1+|P|/|W|)$, which controls sensitivity to temporal span. Adaptation is stabilized by a short-horizon regret functional $\mathcal{R}_u$, which attenuates updates as regret diminishes, promoting convergence. Iterations terminate once successive threshold changes fall below a tolerance or when regret ceases to decrease.

Let $p_{va}$ denote calibrated validation scores. For $\Theta \in T$, the driver-macro harmonic mean (HM) score is defined as:

$$\mathrm{HM_{k,j}}(\Theta; p_{va}(\tau)) = \frac{1}{\left|\mathcal{G}_{k,j}^{\mathrm{v}}\right|} \sum_{g \in \mathcal{G}_{k,j}^{\mathrm{v}}} \frac{2\,\mathrm{Acc}_g(\Theta)\,\mathrm{F1}_g(\Theta)}{\mathrm{Acc}_g(\Theta) + \mathrm{F1}_g(\Theta)} \tag{S6}$$

Here, $\mathcal{G}_{k,j}^{\mathrm{v}}$ is the set of drivers in the validation fold. The performance functional of $\tau$ is then:

$$H_{k,j}(\tau) \coloneqq \max_{\Theta \in \boldsymbol{T}} \mathrm{HM_{k,j}}\left(\Theta;\, p_{\mathrm{va}}(\tau)\right) \tag{S7}$$

The inner-fold adaptation of $\tau$ is thus posed as a one-dimensional, bounded tuning problem:

$$\tau_t \in \arg \max_{\tau \in [\tau_{min}, \tau_{max}]} H_{k,j}(\tau) \tag{S8}$$

evaluated under the calibrated scores $p_{va}(\tau)$. A short-horizon regret functional,

$$\mathcal{R}_u \coloneqq \max_{v \le u} H_{k,j}(\tau_v) - H_{k,j}(\tau_u)\ (\ge 0) \tag{S9}$$

acts as a regularizer: updates are attenuated as $\mathcal{R}_u$ vanishes, ensuring stability. At each candidate τ, labels are recomputed using train-only statistics to ensure fold isolation, so that no validation information enters label construction.

**Inner Tuning and Calibration.** Hyperparameters are tuned by minimizing mean validation negative log-likelihood across inner folds, with **no fixed classification threshold** during tuning. For each candidate, validation probabilities are calibrated (isotonic versus Platt; chosen based on lower validation log-loss), and the operating threshold on validation is

$$\Theta^* = \arg\max_{\Theta \in \mathcal{T}} H\, M_{k,j}(\Theta) \tag{S10}$$

optionally subject to a recall floor $r_{min}$. After completing the $J$ inner folds, the operating thresholds $\{\Theta_j^*\}$ are aggregated by their median, and the calibrator family is selected by majority vote. The resulting configuration is denoted by $\Theta^*$.

**Outer refit, group-aware calibration, and testing**. For each outer fold $k$, the configuration selected from the inner loop (tuned hyperparameters, the frozen feature subset $\hat{F}$, and the operating threshold $\Theta^*$) is refitted on the full outer-training partition $D_k^{\text{tr}}[\hat{F}]$, without resampling. When class weights are used, they are set from outer-train class priors.

To mitigate in-sample bias, probability calibration is recomputed on $D_k^{\text{tr}}$ using group-aware out-of-fold (OOF) predictions: the training drivers are partitioned by identity, OOF probabilities are generated so that each driver is scored only by models that did not train on that driver, and a single monotone calibration map (chosen between isotonic and Platt) is then fitted on the pooled OOF predictions. The calibration family is selected by lower OOF negative log-likelihood, and its parameters are frozen thereafter.

On $D_k^{\text{te}}$, raw probabilities are transformed by this fixed calibrator and classified using the frozen $\Theta^*$. Evaluation is reported at the subject level and then aggregated across outer folds (e.g., macro averages with confidence intervals).

**Feature Selection and Class Imbalance**. (i) *Feature selection*. SHAP is computed on inner-train only within each inner fold; features are retained per fold and aggregated by selection frequency (kept if selected in at least a pre-specified proportion $\pi_0$ of folds or by top-L frequency). The resulting subset $\hat{F}$ is frozen and used unchanged for outer refit and testing. (ii) *Imbalance*. Class imbalance is handled only during inner model selection (e.g., SMOTE with a safe neighbor choice); the outer refit uses no resampling. When applicable, model-native class weights are set from the outer-train priors.

## III. Evaluation Metrics

To assess the performance of the overall framework and individual models, multiple evaluation metrics were employed. These metrics were selected to capture various aspects of classification performance, particularly under class-imbalanced conditions, which are common in behavioral safety applications. The metrics are accuracy, precision, recall, F1 score, Matthews Correlation Coefficient (MCC), Area Under the Precision–Recall Curve (AUC–PR), respectively:

$$\text{Accuracy } = \frac{TP+TN}{TP+TN+FP+FN} \tag{S11}$$

where TP is the number of true positives, FP is the number of false positives, FN is the number of false negatives and TN is the number of true negatives.

$$\text{Precision} = \frac{TP}{TP+FP} \tag{S12}$$

$$\text{Recall} = \frac{TP}{TP+FN} \tag{S13}$$

$$F_1 = \frac{2\cdot\text{Precision}\cdot\text{Recall}}{\text{Precision + Recall}} \tag{S14}$$

- Matthews Correlation Coefficient (MCC): MCC provides a balanced evaluation even in the presence of class imbalance, taking into account all four quadrants of the confusion matrix. It is computed as:

$$\mathrm{MCC} = \frac{TP \cdot TN - FP \cdot FN}{\sqrt{(TP+FP)(TP+FN)(TN+FP)(TN+FN)}} \quad \text{(S15)}$$

- Area Under the Precision–Recall Curve (AUC–PR): AUC–PR quantifies the trade-off between precision and recall at different thresholds. Unlike ROC AUC, AUC–PR is more informative for imbalanced datasets, where the positive class is rare. It is defined as the area under the curve formed by plotting Precision versus Recall.

## IV. Hyperparameter Tuning

Table SI shows the results for model hyperparameter tuning, including the hyperparameters, their search space, the best values identified within the Optuna trials, and a brief description of each. The optimized settings reveal clear domain-specific adaptation patterns, with models tuned on harsh events favoring deeper architectures, lower regularization, and faster convergence, while those trained on speed-weighted headway adopt more regularized, shallower structures with longer convergence times. Such differences reflect the framework's ability to tailor model complexity and learning dynamics to the statistical and temporal characteristics of each safety indicator under dynamic thresholding and regret-based optimization.

## V. Regret Tracking Results

Fig. S1 illustrates the evolution of regret across optimization trials for RF, XGB, and DNN under dynamic threshold adaptation, separately for harsh events (top) and speed-weighted headways (bottom). In the harsh events domain, XGB and DNN began with minimal regret, reflecting effective initial $\tau_t$ calibration. Over time, RF showed a gradual increase in regret, particularly beyond trial 7, indicating that progressively higher $\tau_t$ values made the model more conservative, which in turn reduced its ability to detect genuine critical events and led to higher regret. XGB showed a transient spike around trials 10–12 but quickly stabilised, but quickly stabilised, indicating robust convergence as the empirical Bernstein-based feedback mechanism narrowed the allowable $\tau_t$ adjustments in response to reduced variance in event proportions. Similar to Figure 3, in the harsh events domain the DNN revealed greater volatility across trials, with a standard deviation of 0.089 in HM and a maximum swing magnitude of 0.230, indicating heightened sensitivity to small variations in event proportions. It also completed fewer trials (18) compared to the tree-based models, reflecting the impact of stricter early-pruning criteria and CDSC-driven convergence filtering. In the headway domain, regret values were generally higher across models, reflecting the greater challenge of detecting sparse and temporally constrained headways. RF and DNN showed moderate to high variability across trials, with DNN showing sharper oscillations likely tied to sensitivity in short-term event fluctuations. XGB, contrary to its relatively stable behavior in the harsh event domain, displayed persistent high regret and greater instability in this domain, indicating weaker adaptation to rolling-window, regret-driven threshold updates for proximity-based risk signals.

## VI. Model output analysis

Fig. S2 presents a dual-view of feature importance for harsh event prediction, combining normalized SHAP and tree-based metrics (bar chart) with directional insights from SHAP values (summary plot). Table SII shows the additional questionnaire-derived features visible in the figure and maps each to its cognate category defined in Table I (main manuscript), clarifying how these constructs relate to the core variable taxonomy. Variables prefixed with "ME_" in both Fig. S2 and Fig. S3 correspond to Mobileye ADAS sensor outputs (e.g., ME_Car_speed, ME_AWS_hw_measurement), while GPS-derived measures provide

TABLE SI
MODEL HYPERPARAMETERS WITH OPTUNA SEARCH SPACES AND BEST VALUES

| Model | Hyperparameter | Search Space | Best* | Description |
|---|---|---|---|---|
| RF | n_estimators<br>max_depth<br>min_samples_split<br>min_samples_leaf<br>max_features<br>criterion<br>patience (early stop) | {100, 200, ..., 1000}<br>(step=100)<br>{3, 4, ..., 20}<br>{2, 3, ..., 50}<br>{1, 2, ..., 20}<br>{None, 'sqrt', 'log2'}<br>{'gini', 'entropy', 'log_loss'}<br>{5, ..., 20} | 700 (HA) / 400 (HW)<br>20 (HA) / 14 (HW)<br>3 (HA) / 12 (HW)<br>12 (HA) / 15 (HW)<br>sqrt<br>gini (HA) / log_loss (HW)<br>7 (HA) / 20 (HW) | Number of trees in the forest.<br>Maximum tree depth.<br>Minimum samples to split a node.<br>Minimum samples at a leaf node.<br>Number of features to consider when splitting.<br>Metric for measuring split quality.<br>Number of non-improving iterations before stopping. |
| XGB | n_estimators<br>learning_rate<br>max_depth<br>subsample<br>colsample_bytree<br>gamma<br>reg_lambda<br>reg_alpha | {100, 200, ..., 2000}<br>(step=100)<br>[0.01, 0.1] (log-uniform)<br>{3, 4, ..., 20}<br>[0.5, 0.9]<br>[0.5, 0.9]<br>[0.0, 5.0]<br>[0.1, 10.0]<br>[0.1, 10.0] | 100 (HA) / 1600 (HW)<br>0.067 (HA) / 0.065 (HW)<br>20 (HA) / 3 (HW)<br>0.526 (HA) / 0.669 (HW)<br>0.705 (HA) / 0.686 (HW)<br>4.046 (HA) / 1.839 (HW)<br>9.869 (HA) / 0.494 (HW)<br>9.920 (HA) / 2.879 (HW) | Number of boosting rounds.<br>Step size shrinkage for updates.<br>Maximum tree depth.<br>Fraction of samples per tree.<br>Fraction of features per tree.<br>Minimum loss reduction for further splits.<br>L2 regularization term.<br>L1 regularization term. |
| DNN | units_1<br>activation_1<br>dropout_1<br>l2_reg_1<br>num_layers<br>units_i (i ≥ 2)<br>activation_i<br>dropout_i<br>l2_reg_i<br>lr<br>epochs<br>batch_size<br><br>optimizer<br>loss | {64, 128, ..., 512} (step=64)<br>{'relu', 'tanh', 'leaky_relu'}<br>[0.1, 0.3] (step=0.05)<br>[1e-4, 1e-2] (log-uniform)<br>{1, 2, 3, 4, 5}<br>{32, ..., units_1} (step=32)<br>{'relu', 'tanh', 'leaky_relu'}<br>[0.1, 0.3] (step=0.05)<br>[0.001, 0.05] (step=0.005)<br>[1e-4, 1e-2] (log-uniform)<br>{50, ..., 150}<br>{128, 192, 256, 320, 384,<br>448, 512} (step=64)<br>{'Adam', 'SGD',<br>'RMSprop'}<br>Binary Cross-Entropy<br>(fixed) | 320<br>relu (HA) / tanh (HW)<br>0.150 (HA) / 0.25 (HW)<br>0.021 (HA) / 0.006 (HW)<br>2<br>128 (HA) / 160 (HW)<br>tanh<br>0.150 (HA) / 0.3 (HW)<br>0.046 (HA) / 0.016 (HW)<br>0.004<br>66 (HA) / 82 (HW)<br>448<br><br>Adam<br>Binary Cross-Entropy | Neurons in first hidden layer.<br>Activation for first hidden layer.<br>Dropout rate for first hidden layer.<br>L2 regularization strength for first layer.<br>Total hidden layers.<br>Neurons in later hidden layers.<br>Activation for later layers.<br>Dropout rate for later layers.<br>L2 regularization for later layers.<br>Learning rate.<br>Training iterations over the dataset.<br>Samples per gradient update.<br><br>Optimization algorithm.<br>Loss function for binary classification. |

*Best hyperparameters found for each model (HA = harsh events; HW = headway).

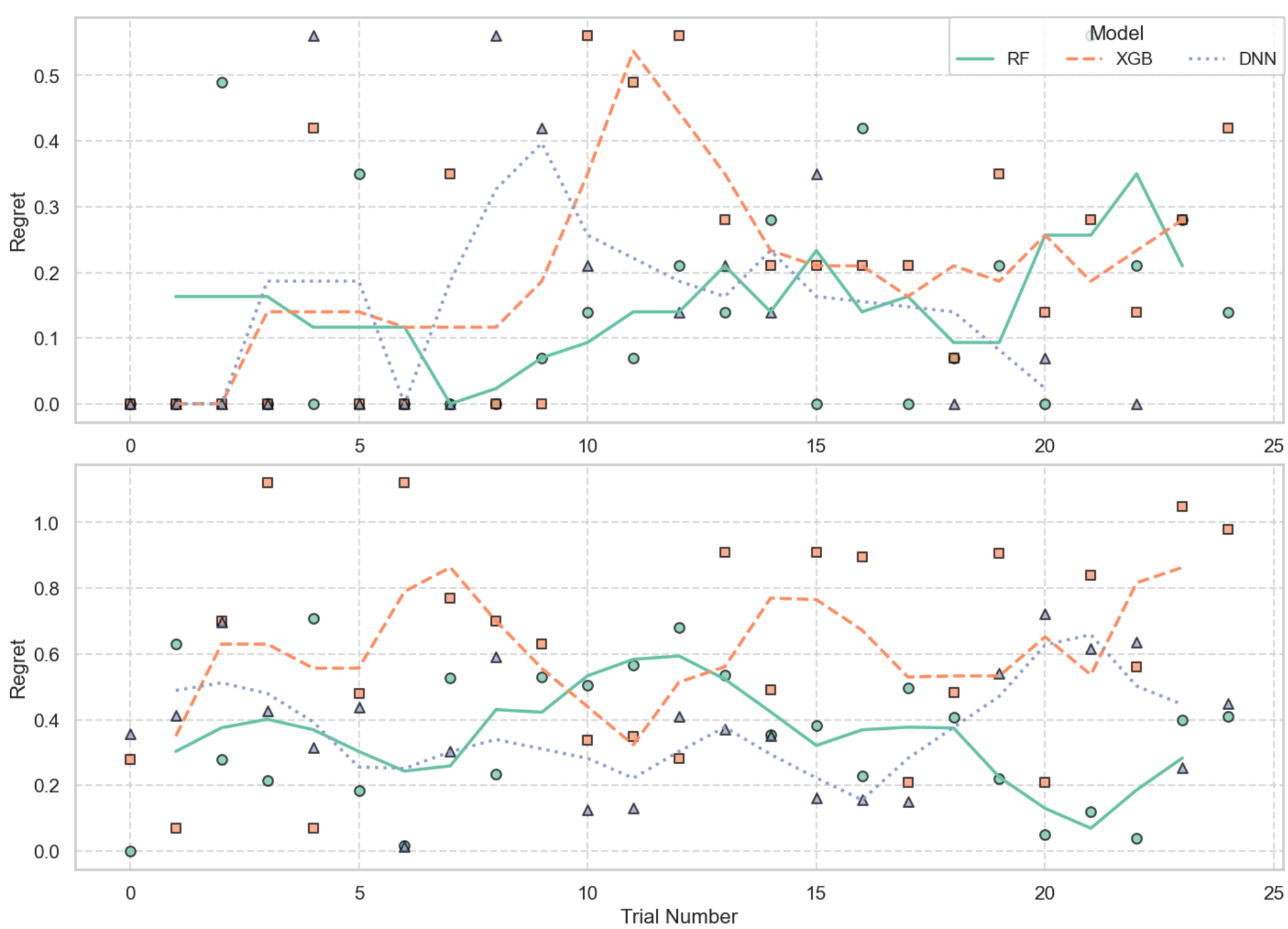

**Fig. SI** Models' regret across optimization trials for harsh events (top) and headway (bottom). Markers are individual-trial regrets; lines are a 3-trial centered moving average per model.

complementary exposure indicators. While SHAP values do not imply causality, they reveal how individual features contribute to the model's predictions. In the summary plot, each point represents a single observation, where color indicates the feature value (blue = low, red = high) and horizontal position reflects its effect on the model output (negative values decrease predicted risk, positive values increase it). The most influential predictors cluster into three conceptual categories: self-regulatory safety attitudes, contextual driving exposure, and kinematic or physiological indicators. Psychometric features such as personal responsibility and perceived control show moderate importance, although they are generally outperformed by direct kinematic indicators. Trip dynamics such as elevated GPS speed, extended trip duration, and high OBD-recorded vehicle speed (ME_Car_speed) show consistently positive SHAP values, linking these patterns to higher predicted risk. Contextual exposure also contributes, with greater rural and motorway driving shares associated with a slight increase in risk, whereas urban driving share shows a model-dependent effect, displaying high tree-based importance but limited SHAP contribution, possibly reflecting complex interactions or redundancy with other features. Physiological measures (e.g., heart rate variability) and trip index variables offer moderate predictive value but exhibit diffuse SHAP distributions, which may capture session-level or fatigue-related variation. Overall, these findings illustrate how the model integrates kinematic, contextual, and attitudinal information to distinguish between harsh and non-harsh driving episodes.

TABLE SII
SUPPLEMENTARY QUESTIONNAIRE-DERIVED VARIABLES MAPPED TO TABLE I CATEGORIES (MAIN MANUSCRIPT)

| Variable | Description | Related Category in Table I (Main Manuscript) |
|---|---|---|
| Impulse Control Score | Agreement with items on ability to restrain risky/impulsive actions. | Driver's attitudes and beliefs |
| Feels in Control When Driving Safely | Belief that one's own behavior determines safe driving. | Driver's attitudes and perceived benefits of safe driving |
| Sees Fast Driving as Safe | Belief that high speeds can be driven safely within personal limits. | Driver's attitudes and beliefs |
| Values Safe Driving | Perceived importance/benefit of driving safely. | Driver's attitudes and perceived benefits of safe driving |
| Accepts Risky Overtake | Attitudinal acceptance of risky/illegal overtaking. | Overtaking |
| Accepts Fast Driving (Medium) | Attitudinal acceptance of some fast driving. | Driver's attitudes and perceived benefits of risky driving |
| Accepts Small Gaps (Moderate) | Belief that smaller following gaps can still be safe. | Driver's attitudes about safe distance / risky behavior |
| Drives Sleepy (Moderate) | Self-reported tendency to drive when sleepy. | Driver's attitudes about drowsiness |
| Drives Tired (Medium) | Self-reported tendency to drive when tired. | Driver's attitudes about drowsiness/fatigue |
| Drives Fast but Careful (Low/Medium) | Self-perception of fast yet careful driving. | Driver's attitudes and perceived benefits of risky driving |
| Friends Value Safety | Social norm: friends consider safe driving important. | Driver's attitudes and perceived benefits of safe driving |
| Colleagues Encourage Safety (Very High) | Social norm: colleagues encourage safe driving. | Driver's attitudes and beliefs |
| Belief in Skill Need (High) | Belief that specific skills are required for safe driving. | Driver's attitudes and beliefs |
| Moral Obligation to Drive Safely (High) | Sense of moral duty to drive safely. | Driver's attitudes and perceived benefits of safe driving |
| Self-Reported Confidence | Self-rated confidence in driving skills and crash avoidance. | Driver's confidence |
| Awareness of Risks (Low) | Self-awareness of taking risks (e.g., speed, overtaking). | Driver's attitudes and beliefs |
| Crash Risk Perception (Low) | Perceived likelihood of being involved in a crash. | Driver's attitudes and beliefs |
| Keeps Safe Distance | Self-reported habit of keeping sufficient following distance. | Driver's attitudes about safe distance |
| Binary Education Level | Highest educational attainment recoded as a binary indicator (≥ university degree = 1; otherwise = 0). | Education |

Fig. S3 presents a dual-view of feature relevance for predicting speed-weighted critical headways, combining normalized SHAP and tree-based importance scores with SHAP summary insights. The predictors with the most contribution span three conceptual domains: trip kinematics, driver characteristics, and safety-related attitudes. Kinematic features such as GPS speed, trip duration, and travelled distance consistently rank highest, with positive SHAP values suggesting that faster and longer trips are associated with higher model-predicted risk of headways, potentially reflecting reduced reaction margins at elevated speeds. Contextual exposure, particularly increased motorway and rural driving shares, also showed positive associations in the model's output, indicating that high-speed environments may challenge safe headway maintenance. Driver-related variables, including age, years of driving, and education level, showed moderate predictive value, though their SHAP effects appear bidirectional, indicating individual variability in how these factors influence the model's predictions. Environmental proxies such as wiper activity (indicative of adverse weather conditions) and ADAS inputs tend to be more active during complex driving situations, and their SHAP contributions plateau at higher levels of activation. The Trip Index feature revealed mixed directional SHAP effects, which may reflect session-level dynamics such as fatigue or trip progression. Although lower in rank, some psychosocial constructs, such as acceptance of small gaps and permissive speeding attitudes, show consistent SHAP patterns that could reflect behavioral risk dispositions as perceived by the model. Collectively, these findings reflect how headways are predicted not solely based on immediate speed or static traits, but as a function of how the model integrates environmental, behavioral, and attitudinal inputs recognizing that these patterns are internal to the model and not necessarily causal in nature.

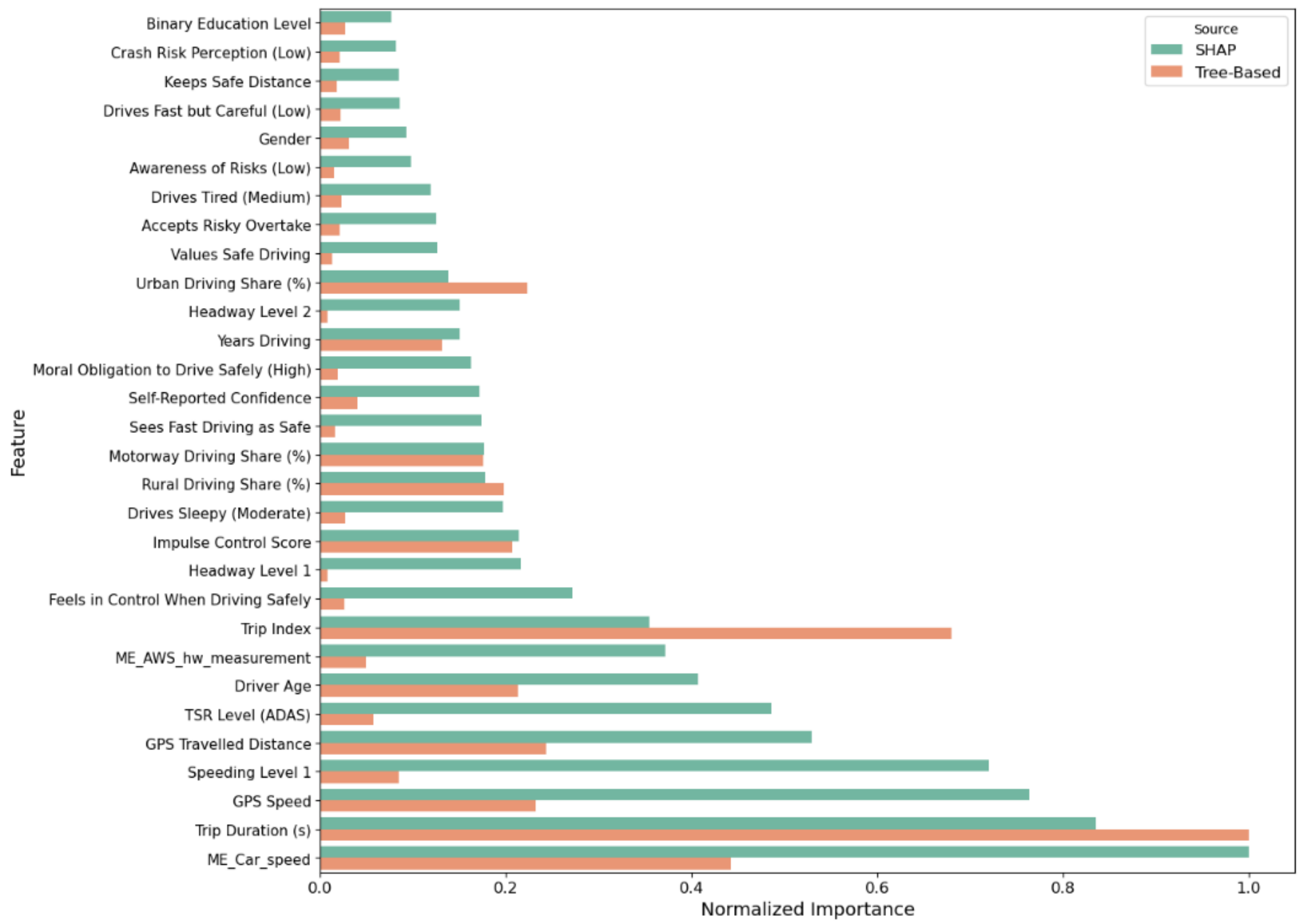

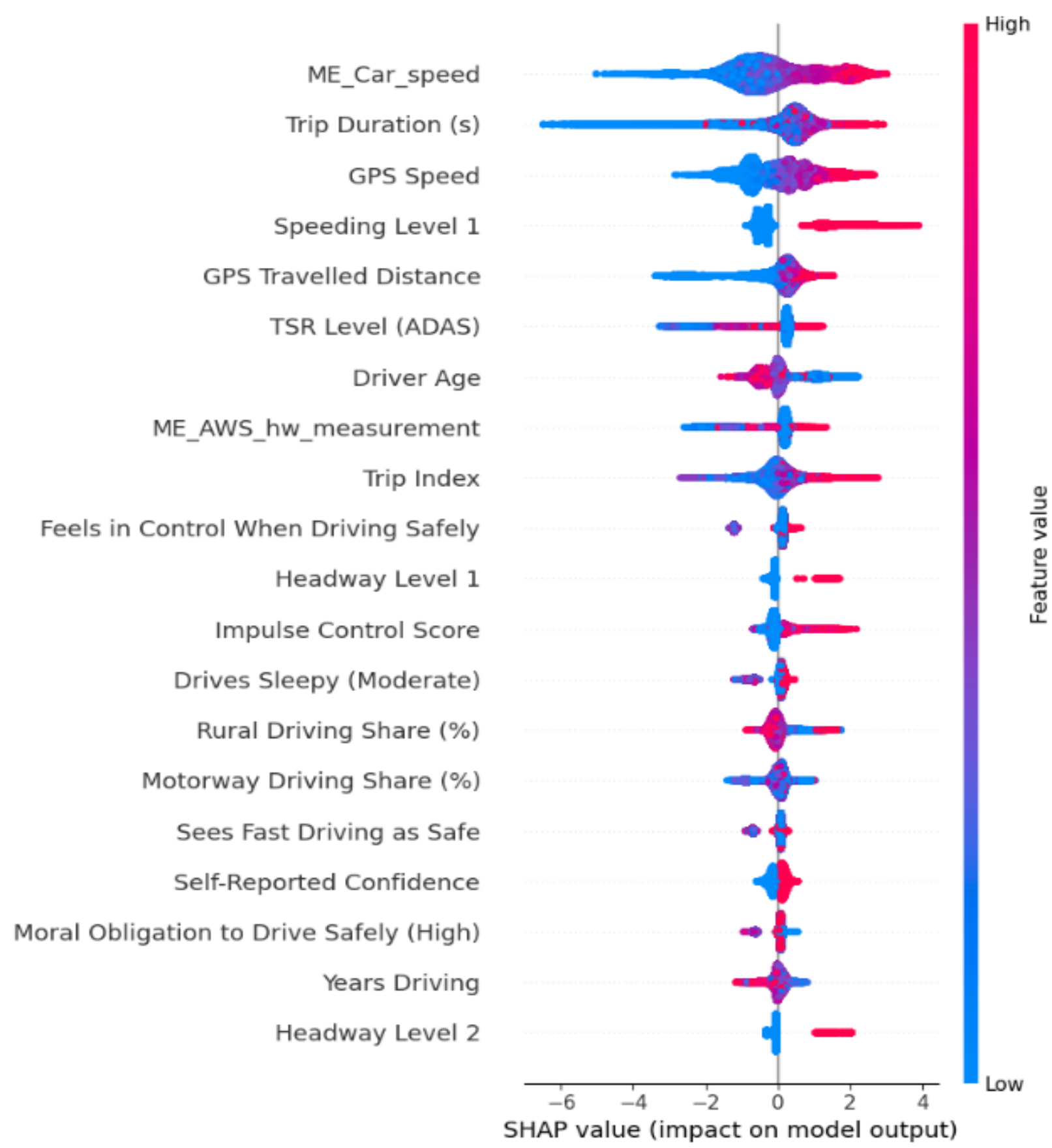

**Fig. S2**. *Dual visualization of feature importance for harsh event prediction, showing the top predictors based on normalized SHAP and tree-based metrics (top) and SHAP value distributions (bottom).*

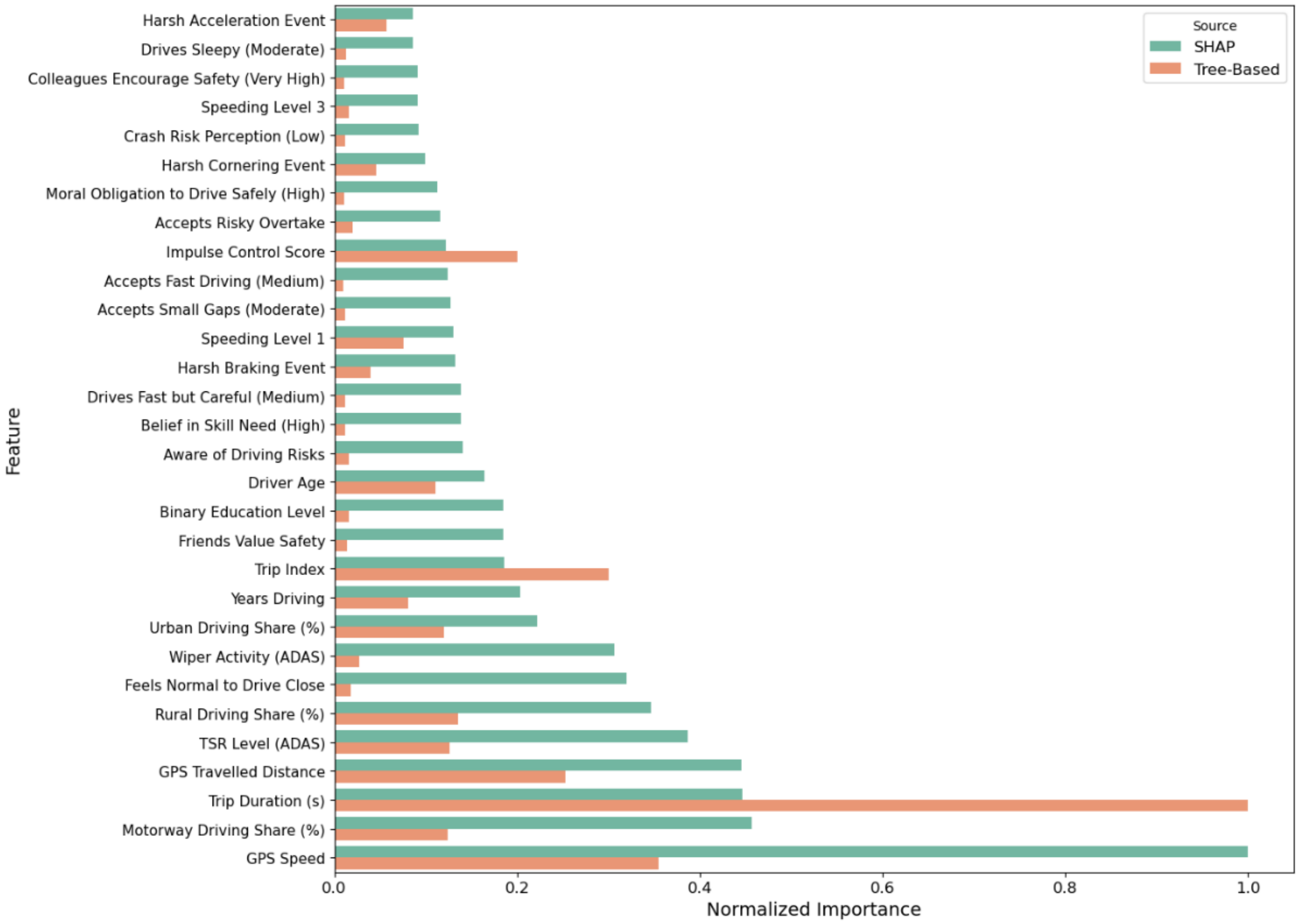

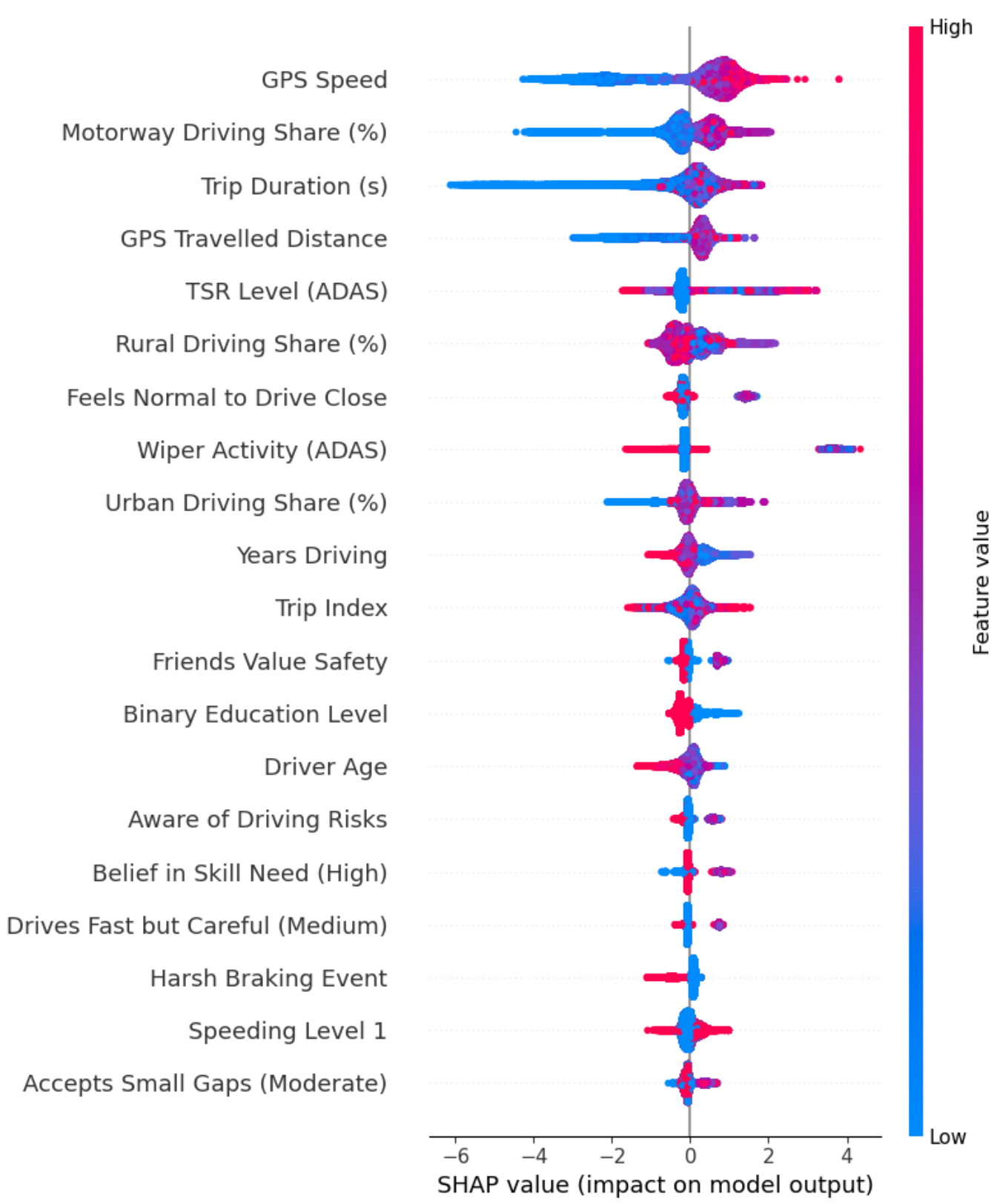

**Fig. S3**. *Dual visualization of feature importance for harsh event prediction, showing the top predictors based on normalized SHAP and tree-based metrics (top) and SHAP value distributions (bottom).*